\documentclass[10pt,twocolumn,letterpaper]{article}

\usepackage{cvpr}
\usepackage{times}
\usepackage{epsfig}
\usepackage{graphicx}
\usepackage{amsmath}
\usepackage{amssymb}
\usepackage{amsthm}
\usepackage{tabularx}
\usepackage{bm, cite, booktabs, enumitem}
\usepackage{alphalph}
\usepackage[algo2e]{algorithm2e} 
\usepackage{algorithm}
\usepackage{algorithmic}
\usepackage[font=small, labelfont=small]{subcaption}
\usepackage[font=small, labelfont=small]{caption}
\usepackage{titling}

\usepackage[pagebackref=true,breaklinks=true,letterpaper=true,colorlinks,bookmarks=false]{hyperref}
\newtheorem{theorem}{Theorem}[section]
\cvprfinalcopy 

\def\cvprPaperID{3913} 

\ifcvprfinal\fi
\begin{document}
\newcolumntype{L}[1]{>{\raggedright\arraybackslash}p{#1}}
\newcolumntype{C}[1]{>{\centering\arraybackslash}p{#1}}
\newcolumntype{R}[1]{>{\raggedleft\arraybackslash}p{#1}}

\makeatletter
\newcommand\nocaption{%
    \renewcommand\p@subfigure{}
    \renewcommand\thesubfigure{\thefigure\alphalph{\value{subfigure}}}
}
\makeatother
\title{\textbf{\Large Fast Single Image Reflection Suppression via Convex Optimization}}
\author{
Yang Yang$^1$, Wenye Ma$^2$, Yin Zheng$^3$, Jian-Feng Cai$^4$, Weiyu Xu$^1$ \\
$^1$University of Iowa, $^2$Tencent, $^3$Tencent AI Lab, $^4$Hong Kong University of Science and Technology\\
\texttt{yy.hz76@gmail.com, wenyema@tencent.com, yinzheng@tencent.com,}\\
\texttt{jfcai@ust.hk, weiyu-xu@uiowa.edu}}

\newcommand\blfootnote[1]{%
  \begingroup
  \renewcommand\thefootnote{}\footnote{#1}%
  \addtocounter{footnote}{-1}%
  \endgroup
}

\setlength{\textfloatsep}{10pt plus 1.0pt minus 2.0pt}
\setlength{\droptitle}{-2em}
\date{}
\maketitle

\begin{abstract}
Removing undesired reflections from images taken through the glass is of great importance in computer vision. It serves as a means to enhance the image quality for aesthetic purposes as well as to preprocess images in machine learning and pattern recognition applications. We propose a convex model to suppress the reflection from a single input image. Our model implies a partial differential equation with gradient thresholding, which is solved efficiently using Discrete Cosine Transform. Extensive experiments on synthetic and real-world images demonstrate that our approach achieves desirable reflection suppression results and dramatically reduces the execution time.
\end{abstract}

\vspace{-0.4cm}
\section{Introduction}\label{sec:intro}\blfootnote{The work of W. Xu was supported in part by Simons Foundation 318608 and in part by NSF DMS-1418737.}
\vspace{-0.5cm}

Images taken through glass usually contain unpleasant reflections.  It is highly desirable if such reflections can be removed. In particular, with the advent of the popularity of portable digital devices such as smartphones and tablets, a lot of such images are taken in everyday life. A fast-response and user-friendly image reflection suppression technology is of great practical significance so that such images can be processed on portable devices in seconds with the best dereflected results produced in real-time according to a user's visual perception.

Given an input reflection-contaminated image $\mathbf{Y}$, traditional approaches that attempt to remove the reflection focus on separating the image into the transmission layer $\mathbf{T}$ (the true background) and the reflection layer $\mathbf{R}$\cite{barrow1978recovering}, i.e., the following assumption is made
\vspace{-0.1cm}
\begin{equation}
    \mathbf{Y} = \mathbf{T} + \mathbf{R},
    \vspace{-0.1cm}
\end{equation}
where $\mathbf{T}$ and $\mathbf{R}$ are unknowns. This problem is highly ill-posed since the number of unknowns is twice the number of conditions. Multiple ways of separation are possible. Different priors and assumptions have been introduced to narrow down the range of valid solutions, despite specific limitations therein. 
\begin{figure}[t]
\nocaption
    \begin{subfigure}[b]{0.235\textwidth}
    \centering
        \includegraphics[width=\textwidth]{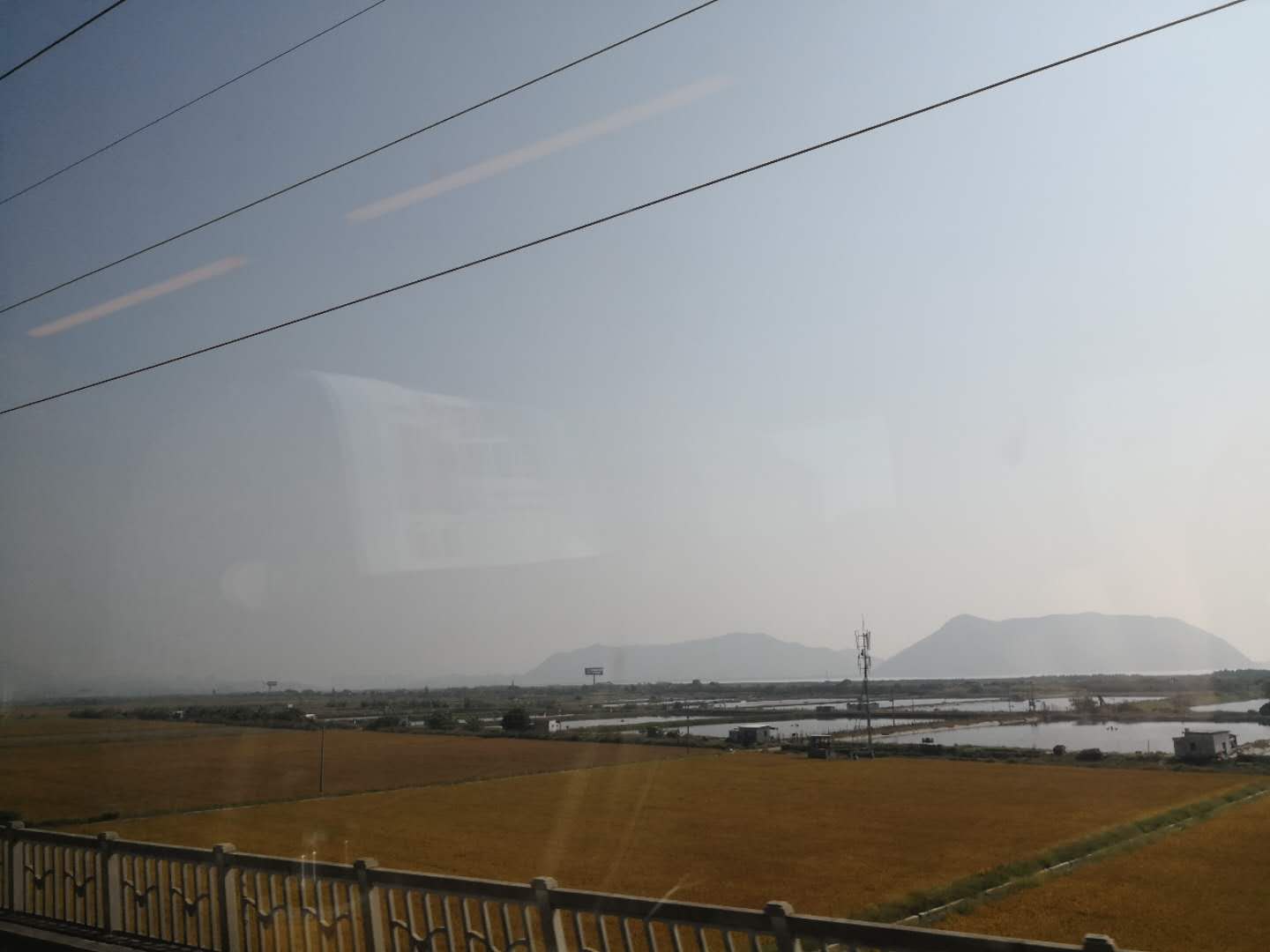}
        \caption{Original Image}
        \label{fig:trainCRH}
    \end{subfigure}
    \hfill    
    \begin{subfigure}[b]{0.235\textwidth}
    \centering
        \includegraphics[width=\textwidth]{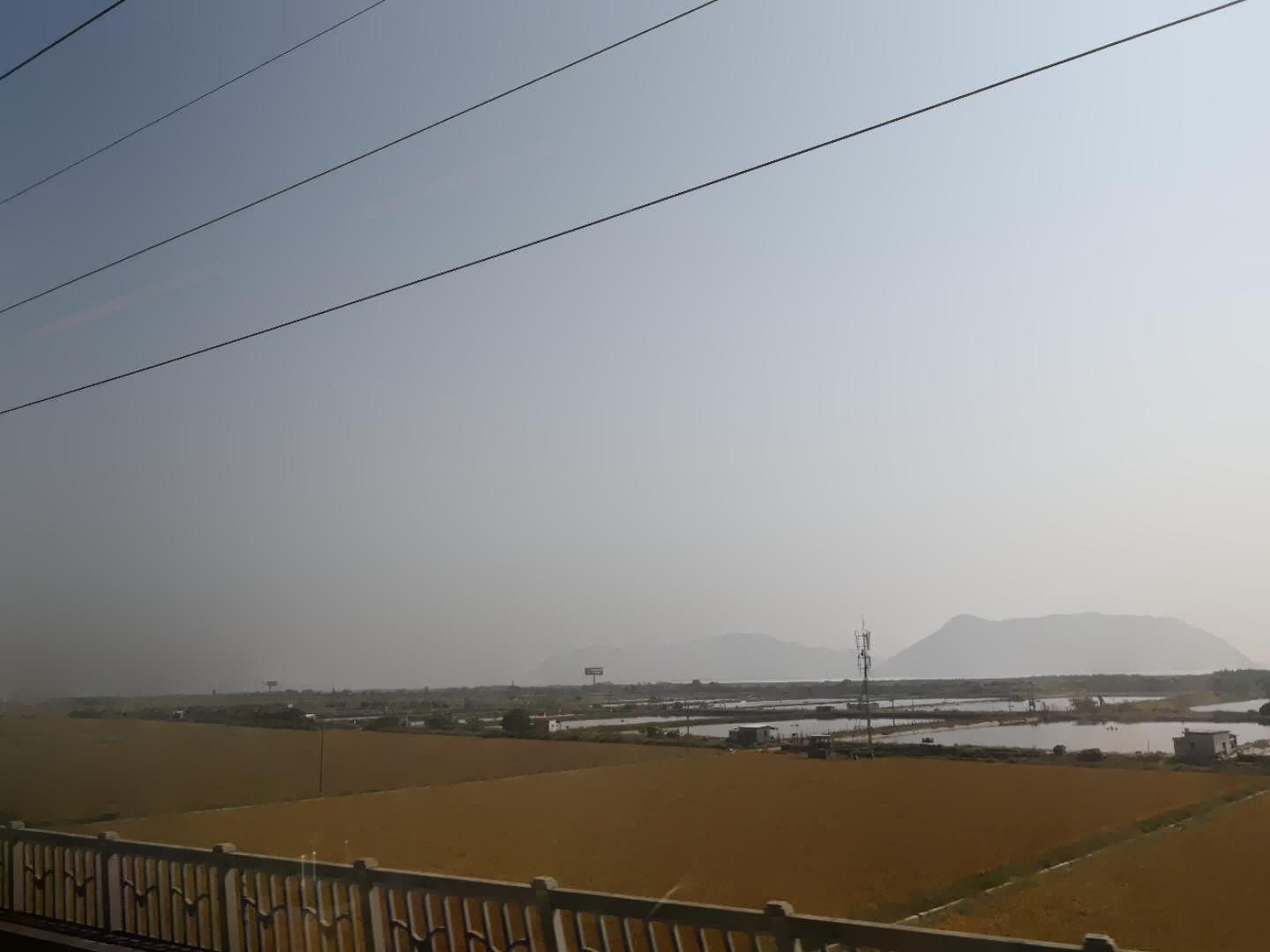}
        \caption{Dereflected Image}
        \label{fig:trainCRHdereflected}
    \end{subfigure}
    \caption{(\ref{fig:trainCRH}): A real-world image taken through the window on a train. Notice the reflection of the seat and the lights in the train. (\ref{fig:trainCRHdereflected}): The result after the reflection suppression by our proposed method. Image size: $1080 \times 1440$. Execution time: 1.15s. \url{https://github.com/yyhz76/reflectSuppress}}
    \label{fig:introduction}
\end{figure}
Instead of \textit{separating} the image into two layers, \textit{suppressing} the reflection in a single input image, as proposed in Arvanitopoulos \etal \cite{arvanitopoulos2017single},  is more practical. In most cases, people are more interested in the transmission layer of an image. Also, perfect layer separation of a single image is in general difficult. The separated layers using existing approaches more or less contain misclassified information, especially when the reflection is sharp and strong, which might yield dark dereflected outputs. This is caused by the removal of a large portion of the energy which concentrates in the reflection layer (See Sec. \ref{sec:experiments}). 

Most image reflection removal approaches so far emphasize the performance in the aspects of the quality of the dereflection. In addition, they can only handle relatively small-sized images and are often computationally inefficient. With the rapid development of portable device technologies, megapixel smartphone images are very common nowadays. Therefore, the efficiency of such methods also needs to be improved to handle large images. We propose an image reflection suppression approach that is highly efficient, which is able to process large smartphone images in seconds, yet can achieve competitive dereflection quality compared to state-of-the-art approaches. Fig.\,\ref{fig:introduction} is an example of our approach applying on a smartphone image.

\subsection{Related Work}
Prior research in image reflection removal can be categorized by the number of input images. One branch relies on multiple input images that are closely related to each other. The other branch only has one image as input.

\subsubsection{Multiple Image Reflection Removal}
The multiple images used for reflection removal are usually related to each other in certain aspects. For example, Schechner \etal\cite{schechner1999polarization}, Farid and Adelson\cite{farid1999separating}, Kong \etal \cite{kong2014physically} separate transmission and reflection layers by taking images of objects at different angles through polarizers. Agrawal \etal\cite{agrawal2005removing} use images taken with and without flash to reduce reflection. Approaches based on different characteristics of fields in transmission and reflection layers are also proposed\cite{gai2008blindly, li2013exploiting, guo2014robust, xue2015computational, sun2016automatic, han2017reflection}. 
Xue \etal\cite{xue2015computational} utilize the difference of motion fields to separate layers. Li and Brown\cite{li2013exploiting} use SIFT-flow to align multiple images and separate layers according to the variation of gradient fields across images. Similarly, Han and Sim\cite{han2017reflection} extend this idea and compute gradient reliability at each pixel and recover the transmission gradients by solving a low-rank matrix completion problem. Reflection removal using multiple images generally achieves better performance than that using a single image since information across images can be exploited to improve layer separation results. However, these approaches usually requires special settings such as images taken from certain angles and locations, or special devices such as polarizers and flashes, which significantly limit their practicality.

\subsubsection{Single Image Reflection Removal}\label{single}
On the other hand, several approaches have also been attempted to remove reflection from a single input image. Although a single input image is more likely to be encountered in everyday life, it is in fact more challenging than multiple image cases due to the lack of additional inter-image information. Existing approaches rely on different prior assumptions on transmission and reflection layers. Levin and Weiss\cite{levin2007user} employ the gradient sparsity prior with user assisted labels to distinguish between layers. Li and Brown\cite{li2014single} exploit the relative smoothness of different layers to separate them using a probabilistic framework. Shih \etal\cite{shih2015reflection} explore the removal of reflection from double-pane glass with ghosting artifacts. Wan \etal\cite{wan2016depth} utilize multi-scale depth of field to classify edges into different layers.  

Instead of separating layers, Arvanitopoulos \etal\cite{arvanitopoulos2017single} propose to suppress the reflection in a single input image using Laplacian-based data fidelity term and gradient sparsity prior, which achieves desirable quality of dereflection but is not quite efficient due to the fact that their model is non-convex and a large number of iterations is needed to achieve desirable result. Other latest methods include deep learning strategies (Fan \etal \cite{fan2017generic}), and nonlocal similar patch search (Wan \etal\cite{wan2017sparsity}). However, either extra network training time or external image datasets are required. 

\subsection{Our Contribution}
In this paper, we propose an approach for single image reflection suppression that achieves desirable performance in terms of both efficiency and dereflection quality. Our contribution is summarized as follows, which contribute to the high efficiency of our approach:
\begin{itemize}[label = \textbullet]
    \item Our proposed model is convex. The solution is guaranteed to be the global optimal of the model. 
    
    \item The optimal solution is in closed form and doesn't rely on iterative algorithms. It is obtained through solving a partial differential equation, which can be done efficiently using Discrete Cosine Transform. 
    
    \item Our method doesn't require any external dataset or training time as in the aforementioned neural network approaches.  
\end{itemize}

\section{Our Proposed Model}\label{sec:theory}
\subsection{Notations}
Throughout the paper, we use bold letters such as $\mathbf{T}$, $\mathbf{Y}$, $\mathbf{K},\bm{f}$ to denote matrices. Plain letters with subscripts $T_{m,n}$ denotes the element of $\mathbf{T}$ at the intersection of the $m$-th row and the $n$-th column. Elementwise multiplication between matrices is denoted by $\circ$ and convolution is denoted by $*$. 

\begin{figure*}[t]
\nocaption
    \centering
    \mbox{
    \begin{subfigure}[b]{0.19\textwidth}
        \includegraphics[width=\textwidth]{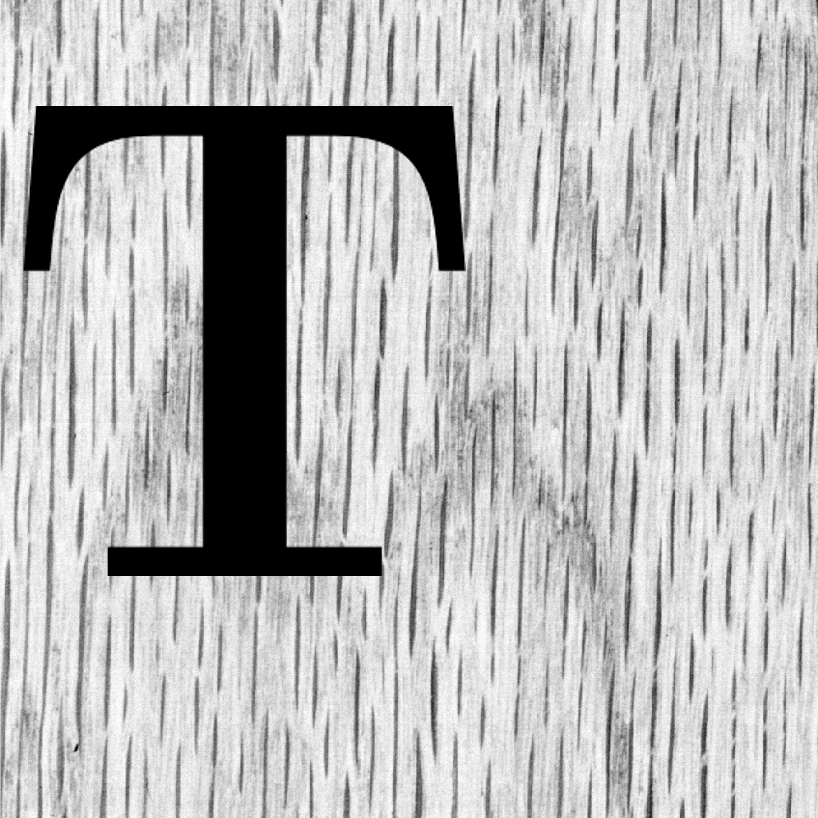}
        \caption{Transmission layer\\{\color{white}a}}
        \label{fig:T}
    \end{subfigure}
    \hfill
    \begin{subfigure}[b]{0.19\textwidth}
        \includegraphics[width=\textwidth]{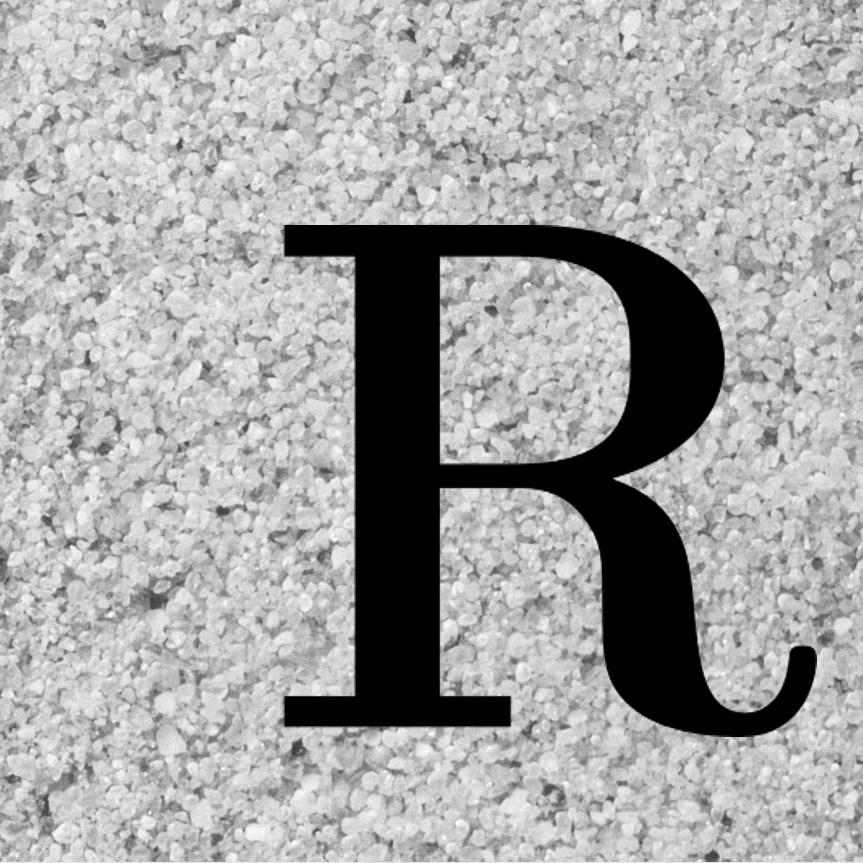}
        \caption{Reflection layer\\{\color{white}a}}
        \label{fig:R}
    \end{subfigure}
    \hfill
    \begin{subfigure}[b]{0.19\textwidth}
        \includegraphics[width=\textwidth]{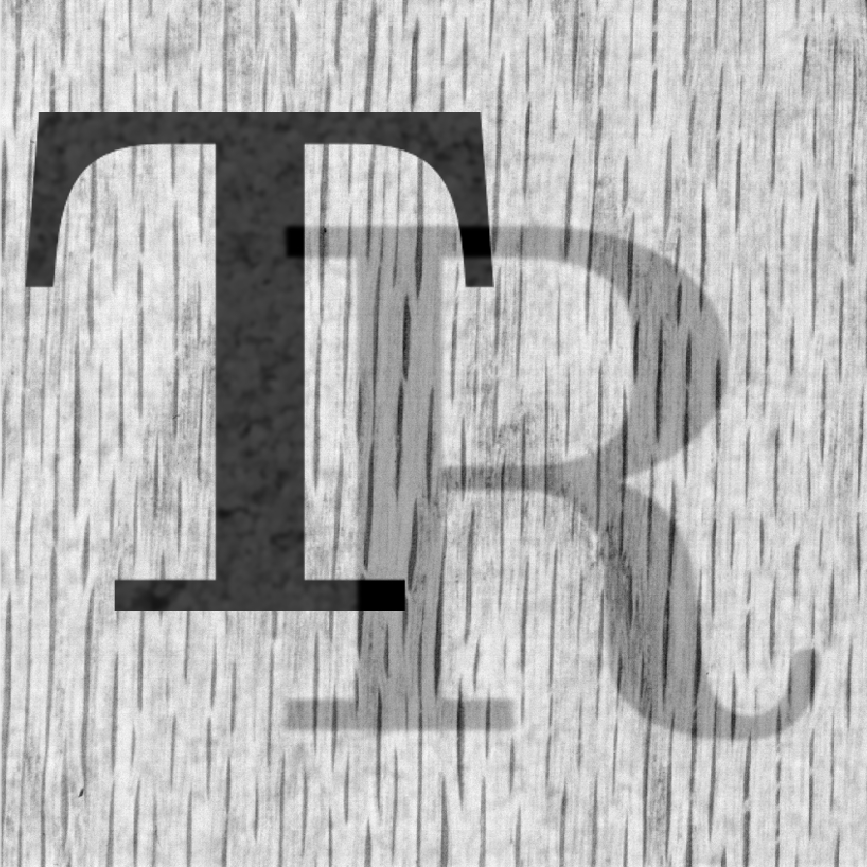}
        \caption{Synthetic blend, \\$w = 0.7, \sigma = 2$}
        \label{fig:blend}
    \end{subfigure}
    \hfill
    \begin{subfigure}[b]{0.19\textwidth}
        \includegraphics[width=\textwidth]{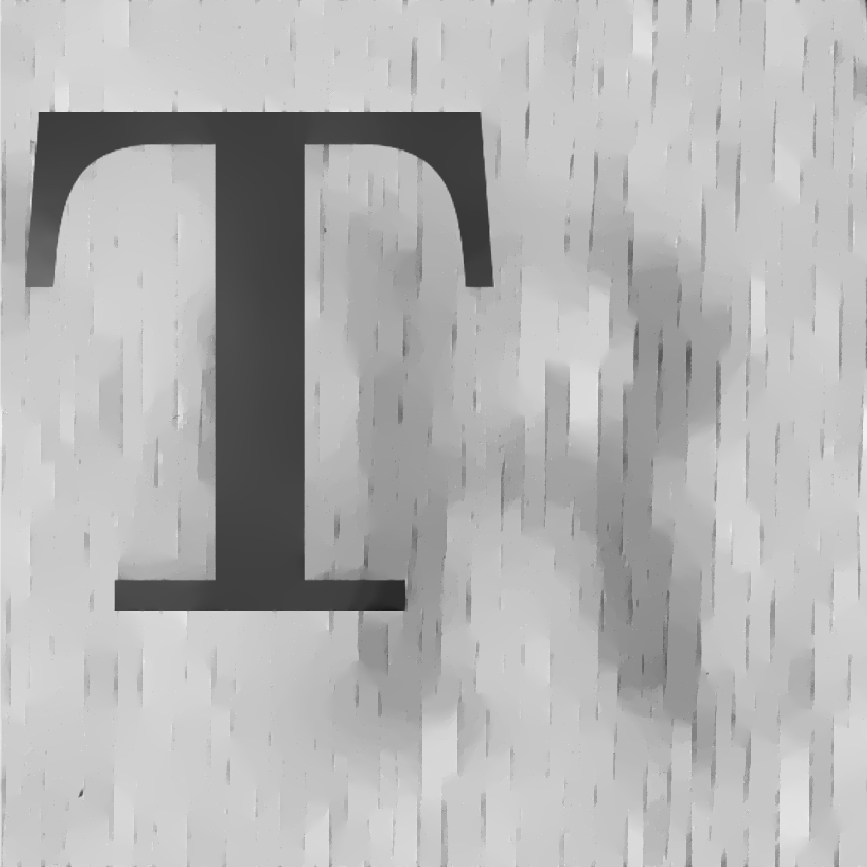}
        \caption{\cite{arvanitopoulos2017single}, $\lambda$ = 0.05.\\Execution time: 382s}
        \label{fig:ar}
    \end{subfigure}
    \hfill 
    \begin{subfigure}[b]{0.19\textwidth}
        \includegraphics[width=\textwidth]{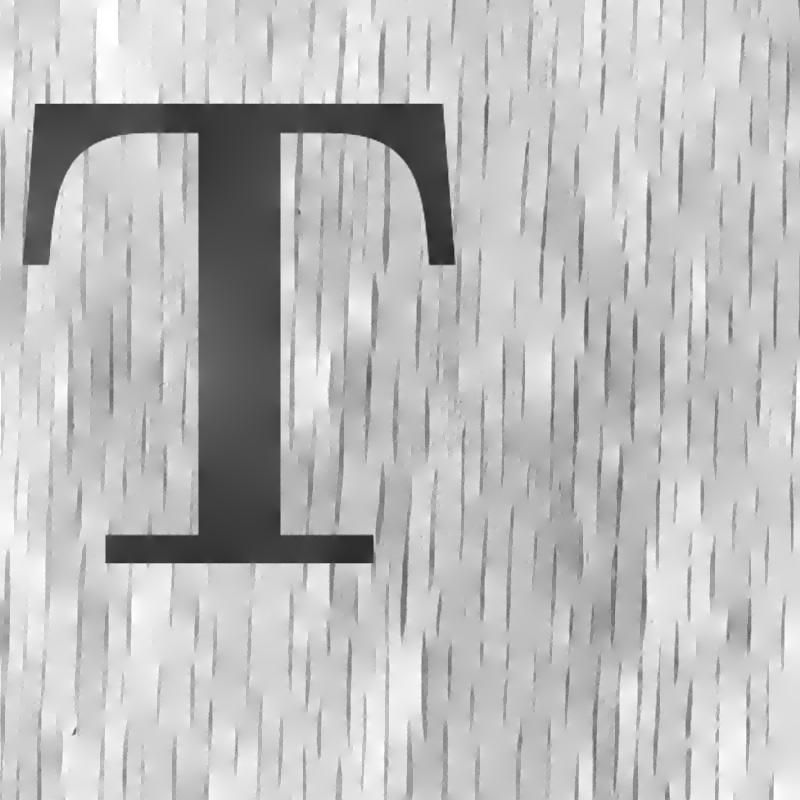}
        \caption{Proposed, $h = 0.11$.\\Execution time: 0.63s}
        \label{fig:proposed}
    \end{subfigure}
    }
    \caption{Comparison of the proposed model with \cite{arvanitopoulos2017single} on a 2D synthetic toy example. The proposed model removes the reflection layer content (i.e., the letter \lq R\rq) more thoroughly. It also retains more transmission layer texture content. The execution time (averaged over 20 repeated runs) of the proposed model is about 600 times faster than \cite{arvanitopoulos2017single}. Image size $800 \times 800$. Texture images from \cite{weber1997usc}.}
    \label{fig:toy example}
\end{figure*}

\subsection{Model Formulation}
Our proposed model relies on the assumption that the camera focuses on the transmission layer (i.e., the objects behind the glass) so that sharp edges appear mostly in this layer. On the other hand, the reflection layer  (i.e., the reflection off the surface of the glass) is less in focus so that edges in this layer are mostly weaker than those in the transmission layer. This is often true in real world scenarios since the distance from the camera to the object in focus is different from that to the glass. We formally express our assumption using the following equation, as mentioned in \cite{arvanitopoulos2017single}: 
\begin{equation}\label{eqn:assumption}
    \mathbf{Y} =  w\mathbf{T} + (1 - w) \mathbf{(\kappa * R)},
\end{equation}
where $\mathbf{Y}$ is the input camera image, $\mathbf{T}$ is the transmission layer and $\mathbf{R}$ is the reflection layer. $w$ is a parameter that measures the weight between the two layers. $\kappa$ is a Gaussian blurring kernel.

Our proposed model is inspired from \cite{arvanitopoulos2017single}, where the original model minimizes the data fidelity term $\|\mathcal{L}(\mathbf{T}) - \mathcal{L}(\mathbf{Y})\|_2^2$ which is the difference on the edges between the output and input images (See Eq.(6) in \cite{arvanitopoulos2017single}). The edge information of an image is obtained by applying the Laplacian operator $\mathcal{L}(\cdot)$ . In addition, an $l_0$ prior of the image gradient  $\|\nabla\mathbf{T}\|_0$ is added to the objective function. It encourages smoothing of the image while maintaining the continuity of large structures. The Laplacian-based data fidelity term better enforces consistency in structures of fine details in the transmission layer compared to a more straightforward data fidelity term\footnote{The data fidelity term $\|\mathbf{T} - \mathbf{Y}\|_2^2$ combined with the $l_0$ prior is used in image smoothing. A detailed discussion can be found in \cite{xu2011image}.} $\|\mathbf{T} - \mathbf{Y}\|_2^2$.
The model in \cite{arvanitopoulos2017single} removes more gradients as the regularization parameter $\lambda$ increases, which is the consequence of using the $l_0$ prior. Essentially, it sets a threshold on the gradients of the input image and removes the gradients whose magnitudes are larger than the given threshold. The \textit{gradient-thresholding} step appears as a closed-form solution in each iteration of their algorithm (See Eq.(12) in \cite{arvanitopoulos2017single}). Similarly, we fuse this idea into our model formulation, but in a different way. Rather than solving the minimization problem and threshold the gradient from the solution, we adopt the idea from \cite{Ma1, Ma2} and put the gradient-thresholding step directly into the objective function. We hence propose the following model:

\begin{equation}\label{eqn:proposed}
\min_{\mathbf{T}}~\frac{1}{2}\|\mathcal{L}(\mathbf{T}) - \text{div}(\delta_h(\mathbf{\nabla Y}))\|_2^2 + \frac{\varepsilon}{2}\|\mathbf{T} - \mathbf{Y}\|^2_2,
\end{equation}
where\begin{align}
\mathcal{L}(\mathbf{Y}) &= \nabla_{xx}(\mathbf{Y}) + \nabla_{yy}(\mathbf{Y}),\\
    \delta_h (X_{i,j}) &= \left\{
    \begin{array}{ll}
    X_{i,j}, & \text{if~} \|X_{i,j}\|_2 \geq h\\\\
    0,              & \text{otherwise}
    \end{array}
    \right. .
\end{align}
The data fidelity term $\|\mathcal{L}(\mathbf{T}) - \text{div}(\delta_h(\mathbf{\nabla Y}))\|_2^2$ imposes the gradient-thresholding step on the input image $\mathbf{Y}$ before taking the divergence of $\mathbf{\nabla Y}$. The gradients whose magnitudes are less than $h$ will become zero. Since the data fidelity term only contains a second order term of the variable $\mathbf{T}$, the second term $\frac{\varepsilon}{2}\|\mathbf{T} - \mathbf{Y}\|^2_2$ is added to guarantee the uniqueness of the solution (see Sec. \ref{solvethemodel} for details), where $\varepsilon$ is taken to be a very small value so as not to affect the performance of the data fidelity term.

Fig.\,\,\ref{fig:toy example} is a toy example demonstrating the effect of our proposed model on synthetic images. We created the transmission layer (Fig. \ref{fig:T}) consisting of a letter \lq T\rq ~and background wooden grain texture. The reflection layer (Fig. \ref{fig:R}) consists of a letter \lq R\rq~and the background sand beach texture. These two layers are then blended (Fig. \ref{fig:blend}) according to Eq.(\ref{eqn:assumption}) with blending weight $w = 0.7$ and the standard deviation of the Gaussian blurring kernel $\kappa$ is set to $\sigma = 2$. We compare the result of \cite{arvanitopoulos2017single} (Fig. \ref{fig:ar}) with our proposed model (Fig. \ref{fig:proposed}).
As can be seen, our proposed model outperforms \cite{arvanitopoulos2017single} both in the quality of dereflection and the execution time. Our proposed method removes the letter \lq R\rq~in the reflection layer while largely preserves the wooden grains in the transmission layer. In contrast, the approach in \cite{arvanitopoulos2017single} doesn't remove the letter \lq R\rq~as thoroughly as ours and a lot more wooden grains are lost. Further increasing the parameter $\lambda$ in \cite{arvanitopoulos2017single} will remove more of the letter \lq R\rq~but at the same time even more wooden grains will be lost as well. In addition, the execution time of our proposed model is much faster than the approach in \cite{arvanitopoulos2017single}.

\subsection{Solving the Model}\label{solvethemodel}
Unlike the model proposed in \cite{arvanitopoulos2017single} which is non-convex due to the presence of the $\|\cdot\|_0$ term, our proposed model (\ref{eqn:proposed}) is convex with respect to the target variable $\mathbf{T}$. Therefore, the optimal solution can be obtained by solving a system of equations, which guarantees the optimality of the solution and contributes to the fast execution time compared to iterative methods that are common among existing approaches (See Sec. \ref{sec:experiments} for details).

The gradient of the objective function (\ref{eqn:proposed}) is given by
\begin{equation}
    \nabla_{\mathbf{T}} = \mathcal{L}\Big(\mathcal{L}(\mathbf{T}) - \text{div}\big(\delta_h(\nabla\mathbf{Y})\big)\Big) + \varepsilon(\mathbf{T} - \mathbf{Y}) .
\end{equation}
Let the gradient be zero, we obtain the following equation
\begin{equation}\label{dct}
    \Big(\mathcal{L}^2 + \varepsilon\Big)\mathbf{T} = \mathcal{L}\Big(\text{div}(\delta_h(\nabla\mathbf{Y}))\Big) + \varepsilon\mathbf{Y} .
\end{equation}
This equation is a variation of 2D Poisson's equation. We associate it with Neumann boundary condition since we assume a mirror extension at the boundary of the image, which implies zero gradient on the boundary. This boundary value problem can hence be solved via \textit{Discrete Cosine Transform} (DCT). Let $\mathcal{F}_c, \mathcal{F}_c^{-1}$ denote the two dimensional DCT and its inverse. We introduce the following result
\begin{theorem}\label{thm1}
The discretization of 2D Poisson's equation 
\begin{equation}\label{poisson}
    \mathcal{L}(\mathbf{T}) = \bm{f}
\end{equation}
with Neumann boundary condition on an $M \times N$ grid is solved by
\begin{equation}\label{solution}
    T_{m,n} = \mathcal{F}_c^{-1}\Bigg(\frac{\left[\mathcal{F}_c(\bm{f})\right]_{m,n}}{K_{m,n}}\Bigg),
\end{equation}
where $\mathbf{T}, \bm{f}, \mathbf{K} \in \mathbb{R}^{M \times N}$. $\displaystyle K_{m,n} = 2\bigg(\cos\left(\frac{m\pi}{M}\right) + \cos\left(\frac{n\pi}{N}\right) - 2\bigg)$. $0\leq m \leq M - 1, 0\leq n \leq N - 1$.
\end{theorem}
See \cite{press2007numerical} for a proof of this conclusion. Essentially it says that after taking DCT, the left side of Eq.(\ref{poisson}) becomes elementwise multiplication, i.e., $\mathcal{F}_c(\mathcal{L}(\mathbf{T})) = \mathbf{K}\circ\mathcal{F}_c(\mathbf{T})$ so the above conclusion follows. It is worth mentioning that the solution (\ref{solution}) has a singularity at $(m, n) = (0, 0)$. To guarantee a unique solution, extra condition (for example, the value at $T_{0,0}$) must be specified beforehand.

We apply Theorem \ref{thm1} to solve Eq.(\ref{dct}). Notice that after taking DCT on both sides, the equation becomes
\begin{equation}
    (\mathbf{K}\circ\mathbf{K} + \varepsilon \mathbf{E})\circ\mathcal{F}_c({\mathbf{T}}) = \mathcal{F}_c(\mathbf{P}),
\end{equation}
where $\mathbf{P}\in \mathbb{R}^{M \times N}$ denotes the right hand side of Eq.(\ref{dct}) and $\mathbf{E} \in \mathbb{R}^{M \times N}$ is a matrix of all 1's. Therefore, the solution to Eq.(\ref{dct}) is 
\begin{equation}\label{sln_proposed}
T_{m,n} = \mathcal{F}_c^{-1}\Bigg(\frac{\left[\mathcal{F}_c(\mathbf{P})\right]_{m,n}}{K_{m,n}^2+\varepsilon}\Bigg),
\end{equation}
where $K_{m,n}$ is the same as in Theorem \ref{thm1}. The uniqueness of the solution is automatically guaranteed because of the presence of $\varepsilon$ in the denominator, which is the consequence of adding the $\frac{\varepsilon}{2}$ term in Eq.({\ref{eqn:proposed}}). Our algorithm is summarized as follows:


\begin{algorithm}[H]
\vspace{0.1cm}
\KwIn{$\mathbf{Y}, h, \varepsilon$}
\begin{algorithmic}
\RETURN $\displaystyle T_{m,n} = \mathcal{F}_c^{-1}
\vspace{0.3cm}
\Bigg(\frac{\left[\mathcal{F}_c(\mathbf{P})\right]_{m,n}}{K_{m,n}^2+\varepsilon}\Bigg)$.\\
\end{algorithmic}
\KwOut{$\mathbf{T}$}
    \caption{{\bf Image Reflection Suppression via Gradient Thresholding and Solving PDE}
    \label{Algorithm}}
\end{algorithm}


\section{Experiments}\label{sec:experiments}All experiments are implemented using MATLAB 2017a on a PC with 8-core Intel i7-8550U 1.80GHz CPU and 16 GB memory.  
We compare our method with state-of-the-art approaches Arvanitopoulos \etal\cite{arvanitopoulos2017single}, Li and Brown\cite{li2014single} and Wan \etal\cite{wan2016depth}.
These approaches are implemented using the original MATLAB source code provided from the authors. These approaches are selected for comparison since \emph{only} a single image is required as the input. Other single image reflection removal approaches mentioned in Sec. \ref{single} either require external image datasets\cite{fan2017generic, wan2017sparsity} or additional conditions (user labels\cite{levin2007user}, double-pane glass and ghosting cues\cite{shih2015reflection}). We use PSNR and SSIM (adopted in \cite{arvanitopoulos2017single}) together with execution time as metrics to evaluate the performance of the selected approaches. The execution times reported throughout this paper are all averaged over 20 repeated runs.

\begin{figure*}[t]
\nocaption
    \begin{center}
    \mbox{
    \begin{subfigure}[b]{0.23\textwidth}
        \includegraphics[width=\textwidth]{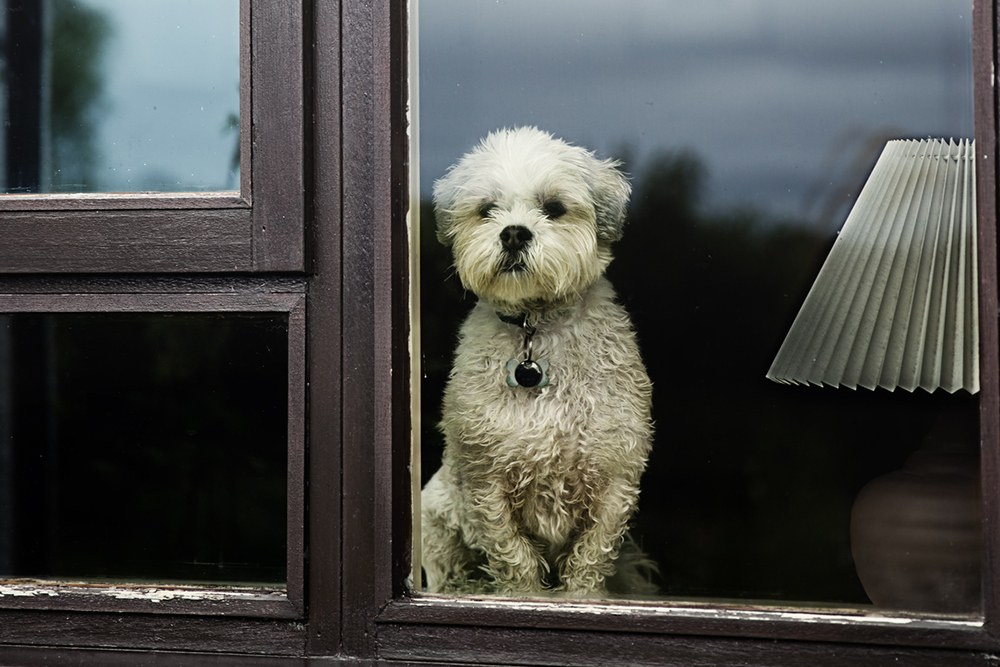}
        \caption{Input}
        \label{fig:gull}
    \end{subfigure}
    ~ 
    \begin{subfigure}[b]{0.23\textwidth}
        \includegraphics[width=\textwidth]{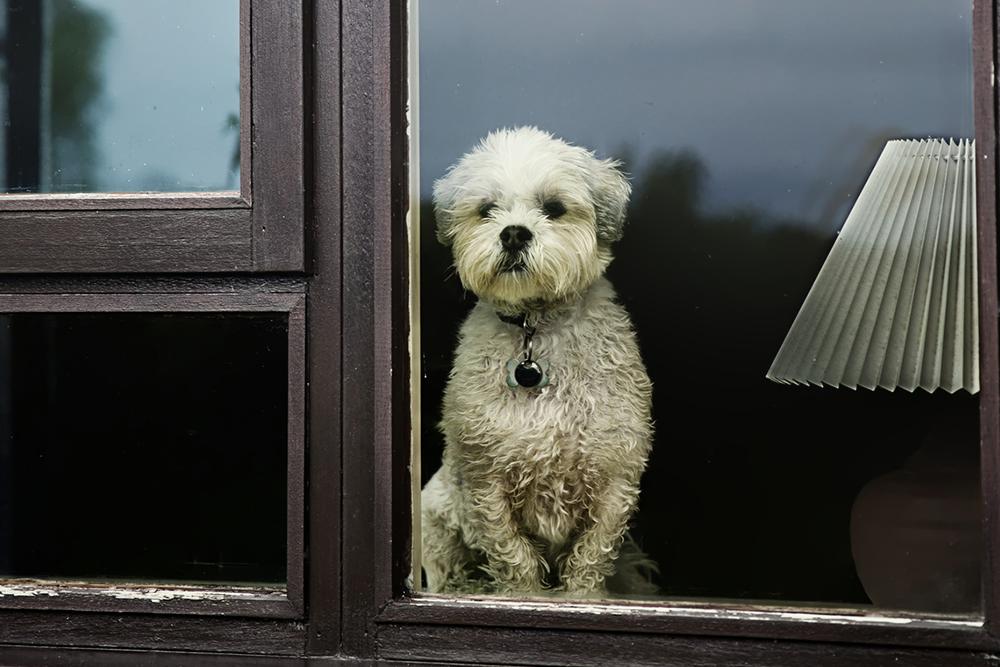}
        \caption{$h = 0.01$}
        \label{fig:tiger}
    \end{subfigure}
    ~ 
    \begin{subfigure}[b]{0.23\textwidth}
        \includegraphics[width=\textwidth]{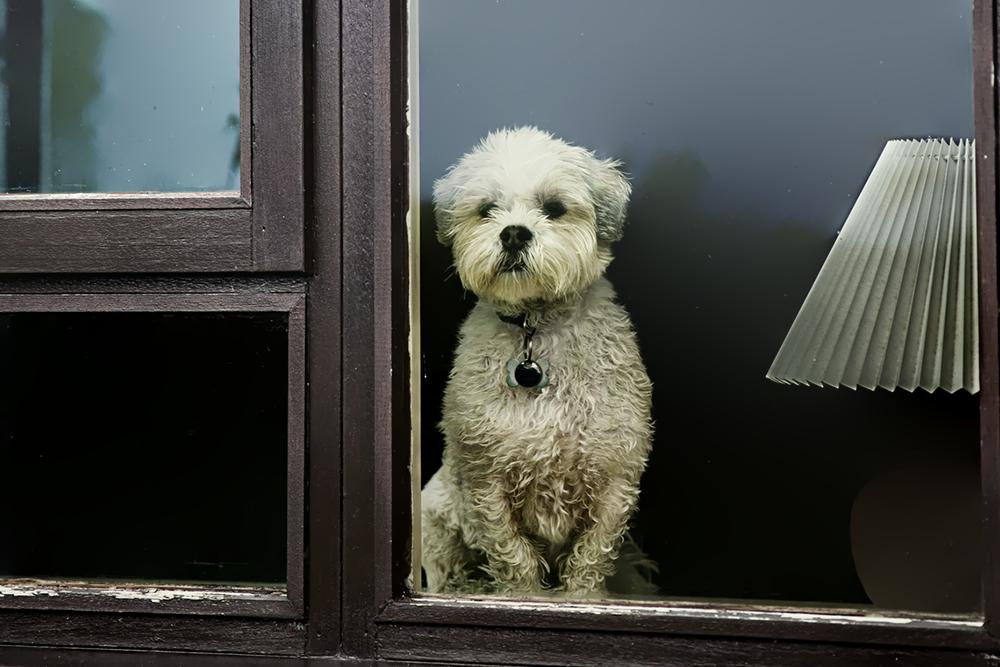}
        \caption{$h = 0.02$}
        \label{fig:mouse}
    \end{subfigure}
    ~
    \begin{subfigure}[b]{0.23\textwidth}
        \includegraphics[width=\textwidth]{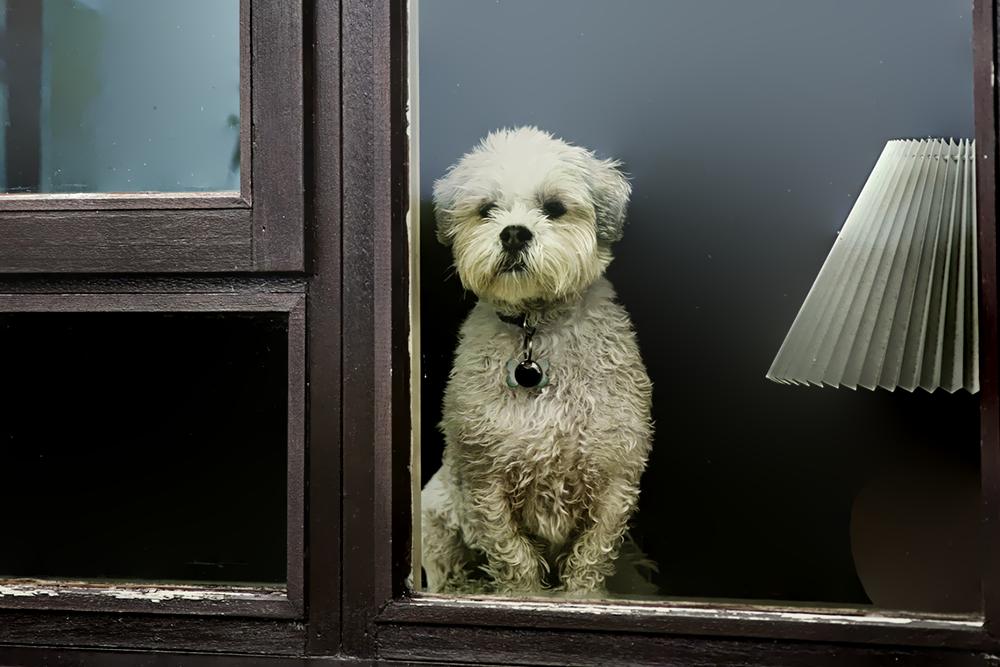}
        \caption{$h = 0.03$}
        \label{fig:mouse}
    \end{subfigure}
    }
    \end{center}
    \vspace{-0.3cm}
    \caption{The effect of increasing the threshold parameter $h$ in the proposed reflection suppression model. Increasing the parameter removes more reflection as well as some details from the transmission layer. Best viewed on screen.}
    \label{parameter change}
\end{figure*}

The parameter $h$ in (\ref{eqn:proposed}) represents the level of the gradient thresholding. The gradients whose magnitudes are less than $h$ will be smoothed out. Fig. \ref{parameter change} shows the effect of increasing $h$. The larger $h$ is, the more reflection components and transmission layer details are removed. Similar to the regularization parameter $\lambda$ in \cite{arvanitopoulos2017single}'s approach, the value of $h$ that produces the best visual result depends on the strength of the reflection in each input image since the best visual result is a balance between the preservation of transmission details and the suppression of reflection. Typically, $h$ values within the interval $[0.01, 0.1]$ yield desirable results. As will be demonstrated below, finding the best parameter $h$ for each image is almost instantaneous. 

\subsection{Synthetic Images}
\begin{figure}[t]
\nocaption
\centering
    \begin{subfigure}[b]{0.23\textwidth}
        \includegraphics[width=\textwidth]{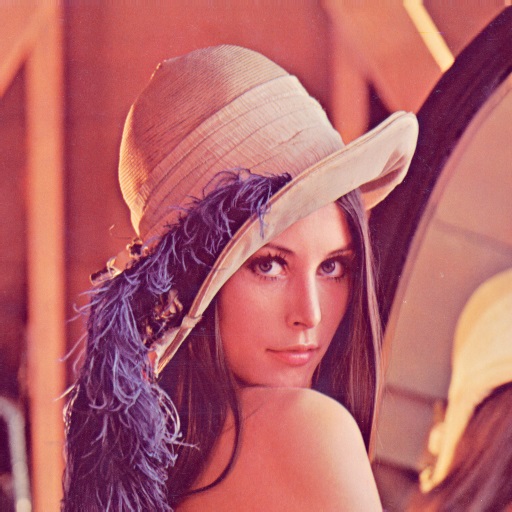}
        \caption{$\mathbf{T_1}$}
    \end{subfigure}
    ~ 
    \begin{subfigure}[b]{0.23\textwidth}
        \includegraphics[width=\textwidth]{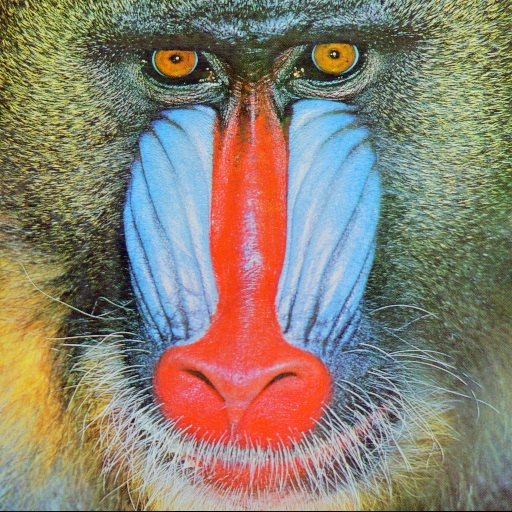}
        \caption{$\mathbf{R_1}$}
    \end{subfigure}
    ~ 
    \begin{subfigure}[b]{0.23\textwidth}
        \includegraphics[width=\textwidth]{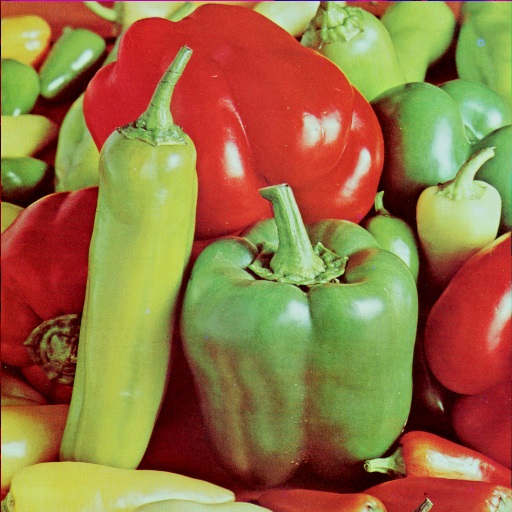}
        \caption{$\mathbf{T_2}$}
    \end{subfigure}
    ~
    \begin{subfigure}[b]{0.23\textwidth}
        \includegraphics[width=\textwidth]{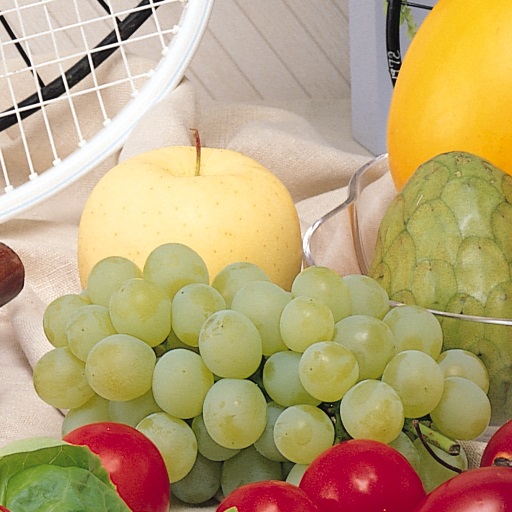}
        \caption{$\mathbf{R_2}$}
    \end{subfigure}
    \caption{Images used as transmission layers ($\mathbf{T_1, T_2}$) and reflection layers ($\mathbf{R_1, R_2}$) for the synthetic experiments. $\mathbf{T_1}$ is blended with $\mathbf{R_1}$. $\mathbf{T_2}$ is blended with $\mathbf{R_2}$.}
    \label{fig:syn}
\end{figure}

We blend two pairs of images of size $512 \times 512$ pixels in Fig.\,\,\ref{fig:syn} according to the assumption (\ref{eqn:assumption}), where $\mathbf{T_i}$ and $\mathbf{R_i}, i = 1,2$ represent transmission and reflection layers, respectively. 
The variance of the Gaussian blurring kernel $\kappa$ is fixed to $\sigma = 4$ and two blending weights $w = 0.7, 0.5$ are used. For parameters in other models, we use the default values as reported in their papers ($\lambda = 100$ in \cite{li2014single}, $\lambda = 0.4$ in \cite{wan2016depth}, $\lambda = 0.002$ in \cite{arvanitopoulos2017single}). In our proposed model, we fix $h = 0.03$ and $\varepsilon = 10^{-6}$.

\begin{figure*}
\nocaption
    \centering
    \begin{subfigure}[b]{0.19\textwidth}
        \includegraphics[width=\textwidth]{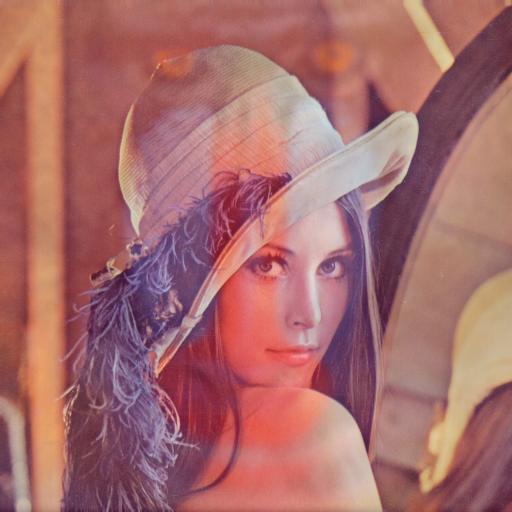}
        \caption{\makebox[1.7cm][l]{$\mathbf{T_1  +  R_1}, w =  0.7$}}
        \label{fig:lena_baboon_07}
    \end{subfigure}
    \begin{subfigure}[b]{0.19\textwidth}
        \includegraphics[width=\textwidth]{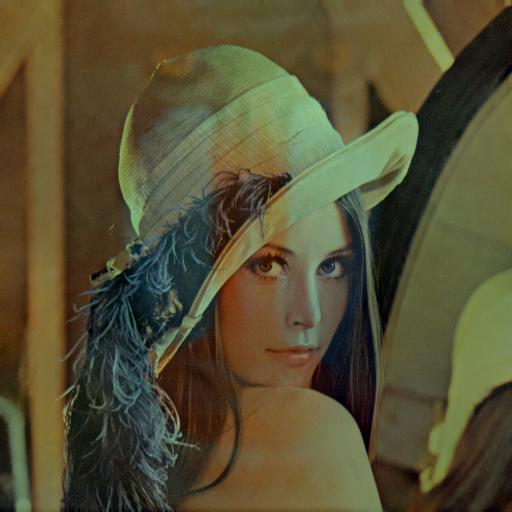}
        \caption{\cite{li2014single}}
    \end{subfigure}
    \begin{subfigure}[b]{0.19\textwidth}
        \includegraphics[width=\textwidth]{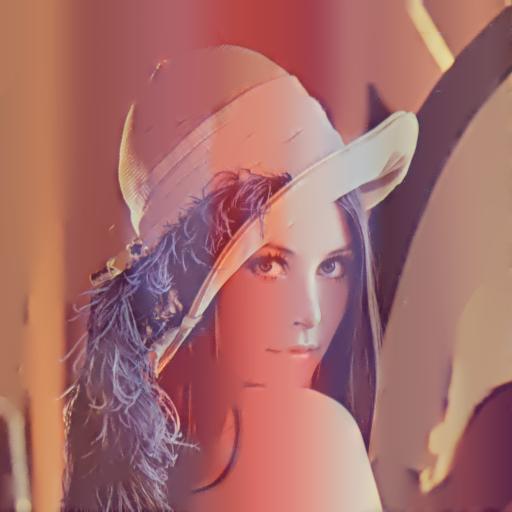}
        \caption{\cite{wan2016depth}}
    \end{subfigure}
    \begin{subfigure}[b]{0.19\textwidth}
        \includegraphics[width=\textwidth]{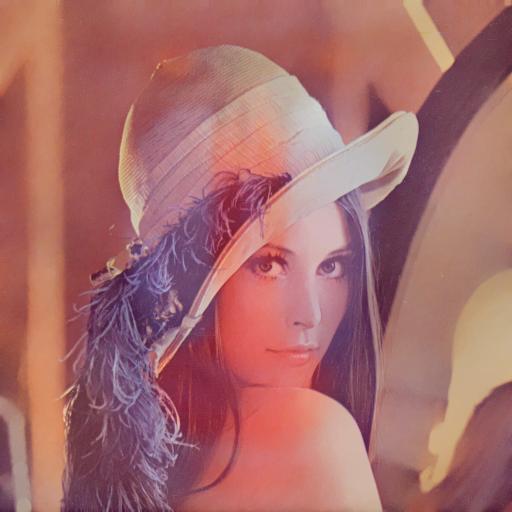}
        \caption{\cite{arvanitopoulos2017single}}
    \end{subfigure}
    \begin{subfigure}[b]{0.19\textwidth}
        \includegraphics[width=\textwidth]{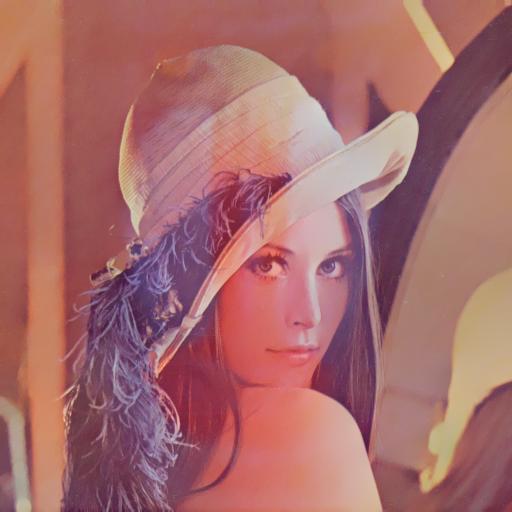}
        \caption{Proposed}
    \end{subfigure} \\
    
    
    \begin{subfigure}[b]{0.19\textwidth}
        \includegraphics[width=\textwidth]{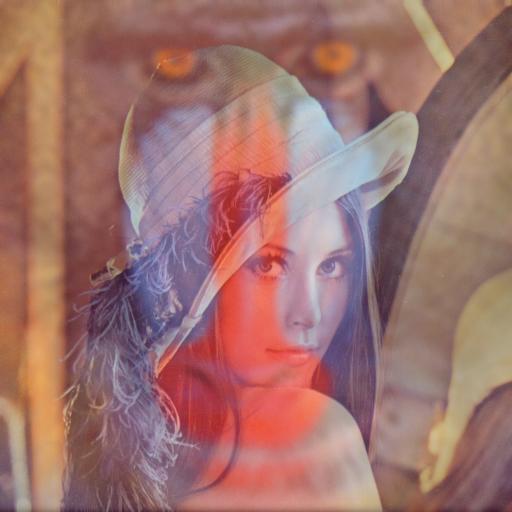}
        \caption{\makebox[1.7cm][l]{$\mathbf{T_1 + R_1}, w = 0.5$}}
        \label{fig:lena_baboon_05}
    \end{subfigure}
    \begin{subfigure}[b]{0.19\textwidth}
        \includegraphics[width=\textwidth]{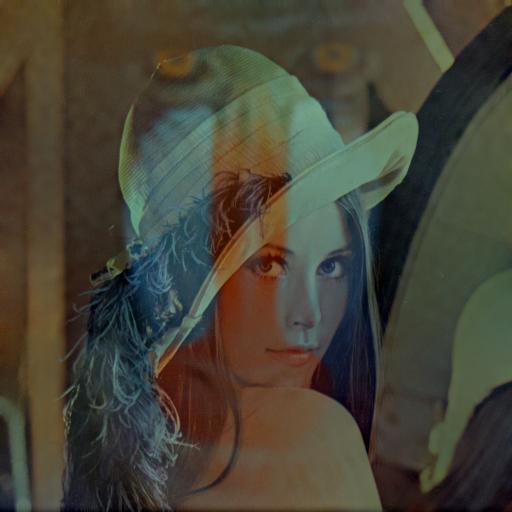}
        \caption{\cite{li2014single}}
    \end{subfigure}
    \begin{subfigure}[b]{0.19\textwidth}
        \includegraphics[width=\textwidth]{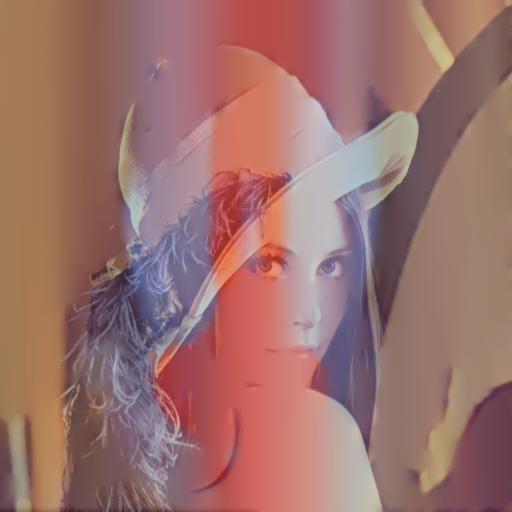}
        \caption{\cite{wan2016depth}}
        \label{fig:lena_wan_05}
    \end{subfigure}
    \begin{subfigure}[b]{0.19\textwidth}
        \includegraphics[width=\textwidth]{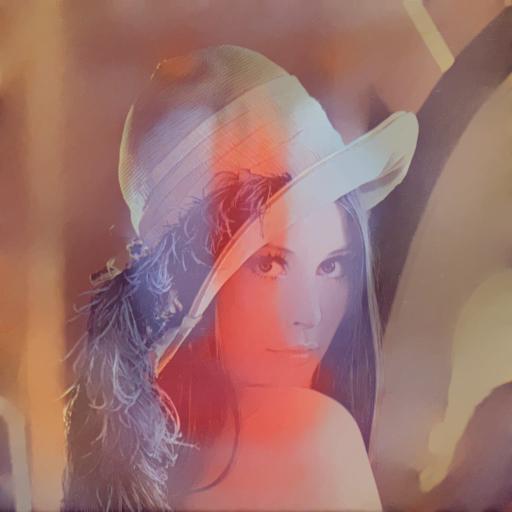}
        \caption{\cite{arvanitopoulos2017single}}
    \end{subfigure}
    \begin{subfigure}[b]{0.19\textwidth}
        \includegraphics[width=\textwidth]{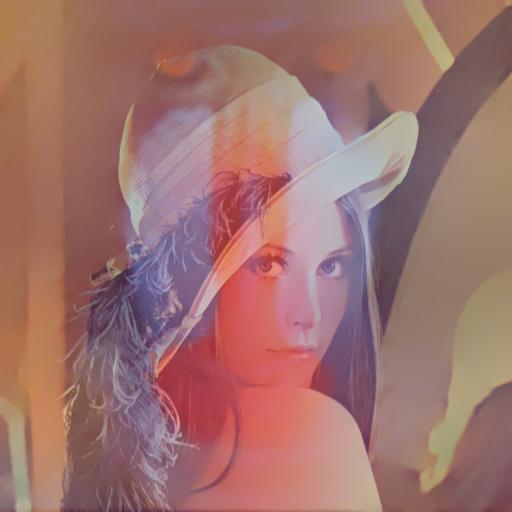}
        \caption{Proposed}
    \end{subfigure} \\
    
    \begin{subfigure}[b]{0.19\textwidth}
        \includegraphics[width=\textwidth]{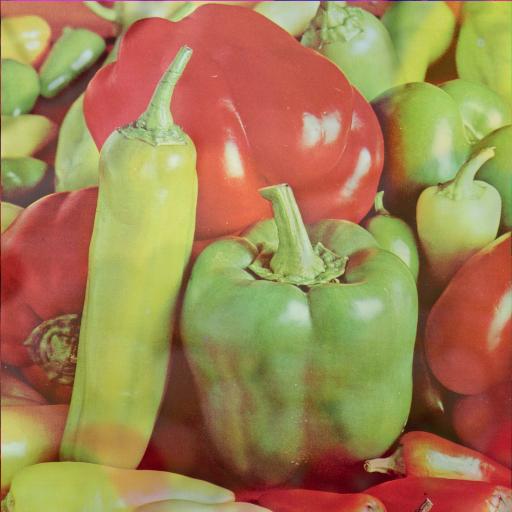}
        \caption{\makebox[1.7cm][l]{$\mathbf{T_2 + R_2}, w = 0.7$}}
        \label{fig:peppers_fruits_07}
    \end{subfigure}
    \begin{subfigure}[b]{0.19\textwidth}
        \includegraphics[width=\textwidth]{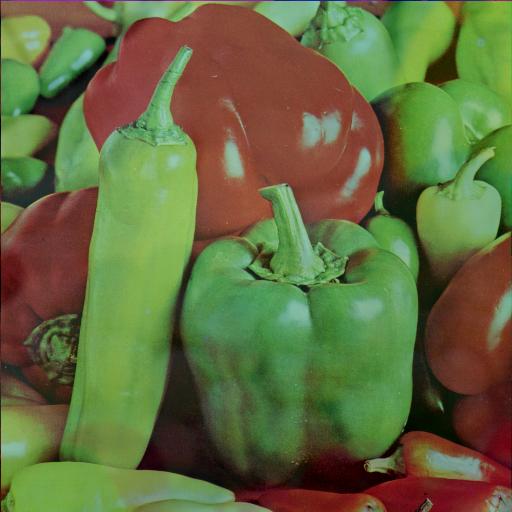}
        \caption{\cite{li2014single}}
    \end{subfigure}
    \begin{subfigure}[b]{0.19\textwidth}
        \includegraphics[width=\textwidth]{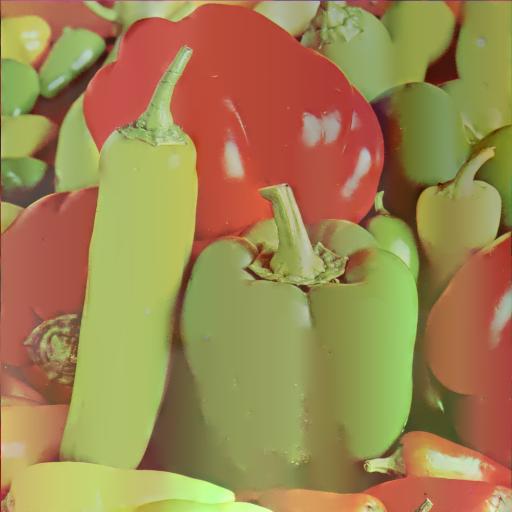}
        \caption{\cite{wan2016depth}}
    \end{subfigure}
    \begin{subfigure}[b]{0.19\textwidth}
        \includegraphics[width=\textwidth]{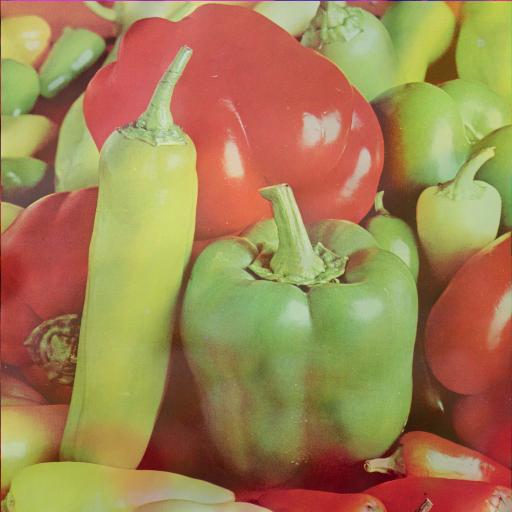}
        \caption{\cite{arvanitopoulos2017single}}
    \end{subfigure}
    \begin{subfigure}[b]{0.19\textwidth}
        \includegraphics[width=\textwidth]{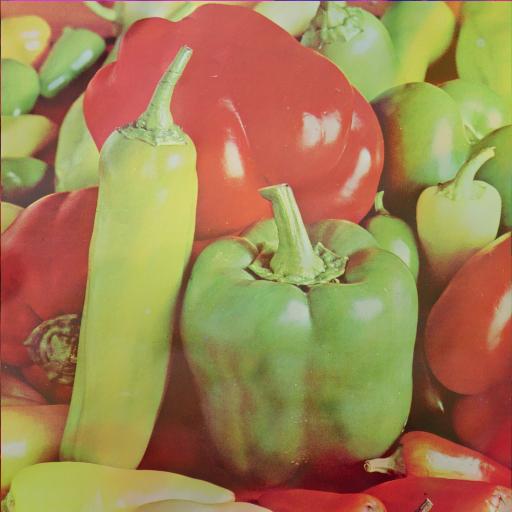}
        \caption{Proposed}
    \end{subfigure} \\
    
%
    \begin{subfigure}[b]{0.19\textwidth}
        \includegraphics[width=\textwidth]{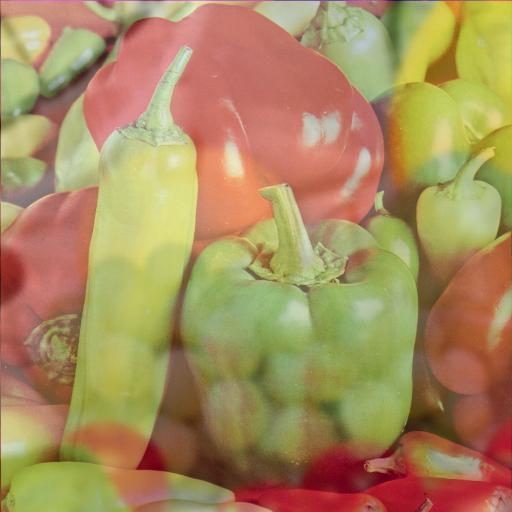}
        \caption{\makebox[1.7cm][l]{$\mathbf{T_2 + R_2}, w = 0.5$}}
        \label{fig:peppers_fruits_05}
    \end{subfigure}
    \begin{subfigure}[b]{0.19\textwidth}
        \includegraphics[width=\textwidth]{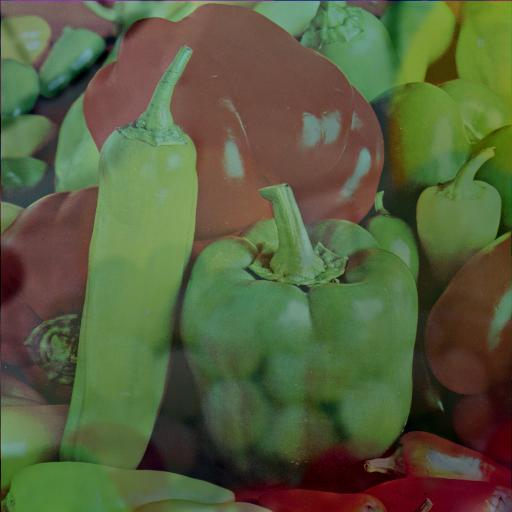}
        \caption{\cite{li2014single}}
    \end{subfigure}
    \begin{subfigure}[b]{0.19\textwidth}
        \includegraphics[width=\textwidth]{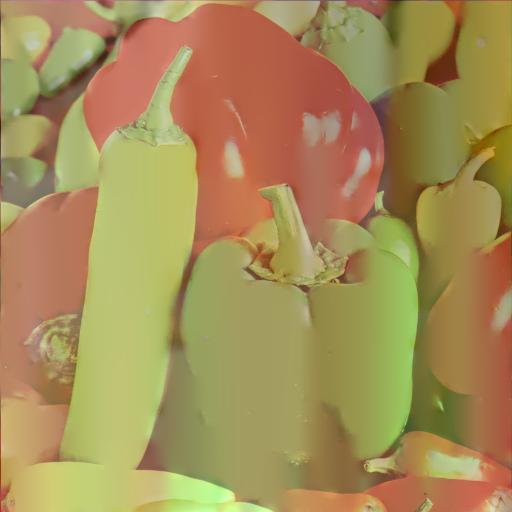}
        \caption{\cite{wan2016depth}}
        \label{fig:pepper_wan_05}
    \end{subfigure}
    \begin{subfigure}[b]{0.19\textwidth}
        \includegraphics[width=\textwidth]{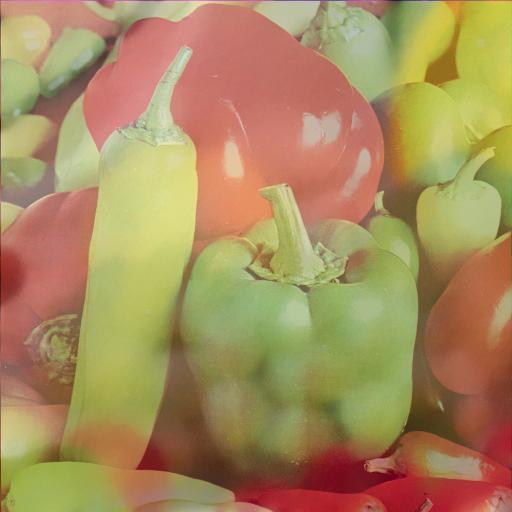}
        \caption{\cite{arvanitopoulos2017single}}
    \end{subfigure}
    \begin{subfigure}[b]{0.19\textwidth}
        \includegraphics[width=\textwidth]{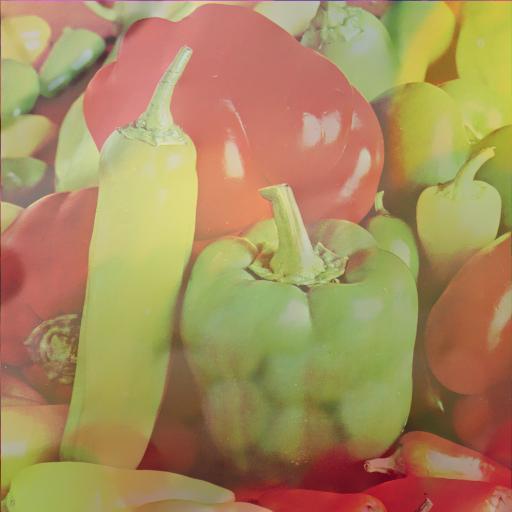}
        \caption{Proposed}
    \end{subfigure}
    \caption{Comparison of reflection suppression on synthetic images. Column 1: Blended images. Column 2: Li and Brown\cite{li2014single}'s results. Column 3: Wan \etal\cite{wan2016depth}'s results. Column 4: Arvanitopoulos\cite{arvanitopoulos2017single}'s results. Column 5: our proposed results. Best viewed on screen.}
    \label{synthetic comparison}
\end{figure*}

The images before and after the reflection suppression are demonstrated in Fig. \ref{synthetic comparison}. The method of Li and Brown\cite{li2014single} tends to produce dark images with false colors. This is partially due to the fact that the energy from the reflection layer accounts for a large portion in our synthetic images. Removing the reflection ends up with significant energy loss and hence produces dark outputs.  The method of Wan \etal\cite{wan2016depth} removes most of the reflection but oversmoothes transmission layer details (For example, top edge of Lena's hat in the mirror, bottom edge of the green pepper, especially in $w = 0.5$ cases (See Fig. \ref{fig:lena_wan_05} and Fig. \ref{fig:pepper_wan_05})). Arvanitopoulos \etal's approach\cite{arvanitopoulos2017single} produces outputs that are the closest to our proposed method. However, as shown in Table \ref{PSNRSSIM} and \ref{time synthetic}, our outputs achieve better performance in terms of PSNR, SSIM and execution time in all cases. Particularly, notice that the execution time of our method outperforms all the others by a significant margin.

\subsection{Real-World Images}
\normalsize
The size of the real-world images used here are $1080 \times 1440$ pixels. We captured these images directly using smartphone. Default parameter settings are used in the method of Li and Brown\cite{li2014single}. As for the method of Arvanitopoulos \etal\cite{arvanitopoulos2017single}, we tune the regularization parameter $\lambda$ for each input image to get the best visual result since the outcome is much more sensitive to parameter tuning compared to Li and Brown's approach. In our proposed model (\ref{eqn:proposed}), the parameter $h$ is tuned for each input image for the same reason. However, parameter tuning in our model is almost instantaneous, which will be demonstrated below. The parameter $\varepsilon$ is empirically fixed to $10^{-8}$.

\begin{figure*}[t]
\nocaption
    \centering
    \begin{subfigure}[b]{0.23\textwidth}
        \includegraphics[width=\textwidth]{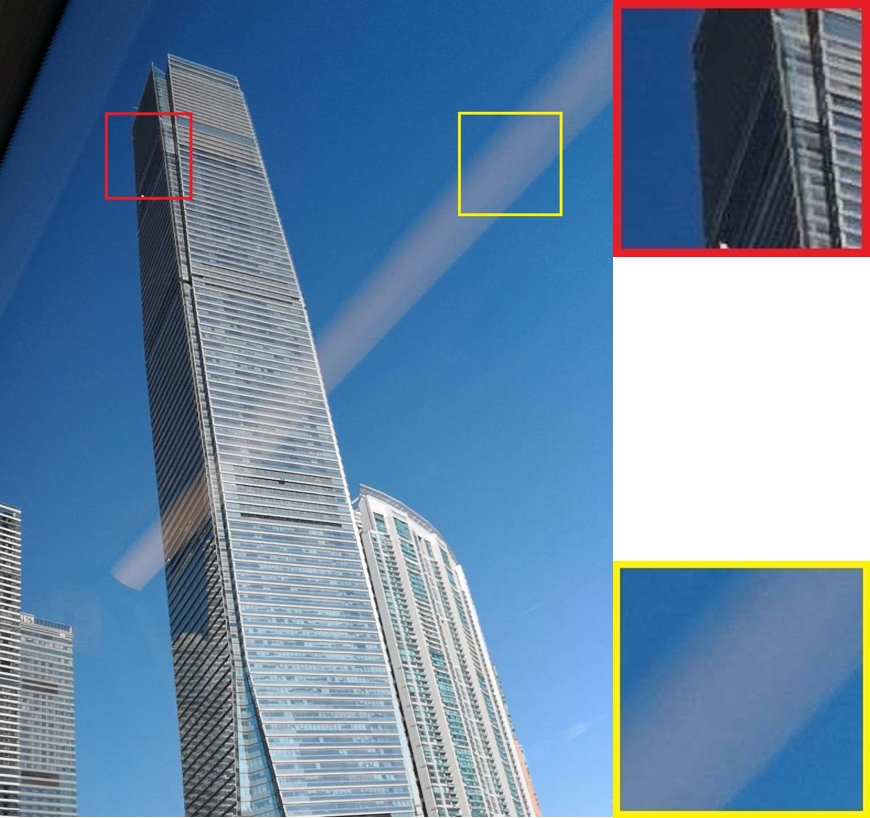}
        \caption{Input 1\\~}
        \label{fig:gull}
    \end{subfigure}
    ~ 
    \begin{subfigure}[b]{0.23\textwidth}
        \includegraphics[width=\textwidth]{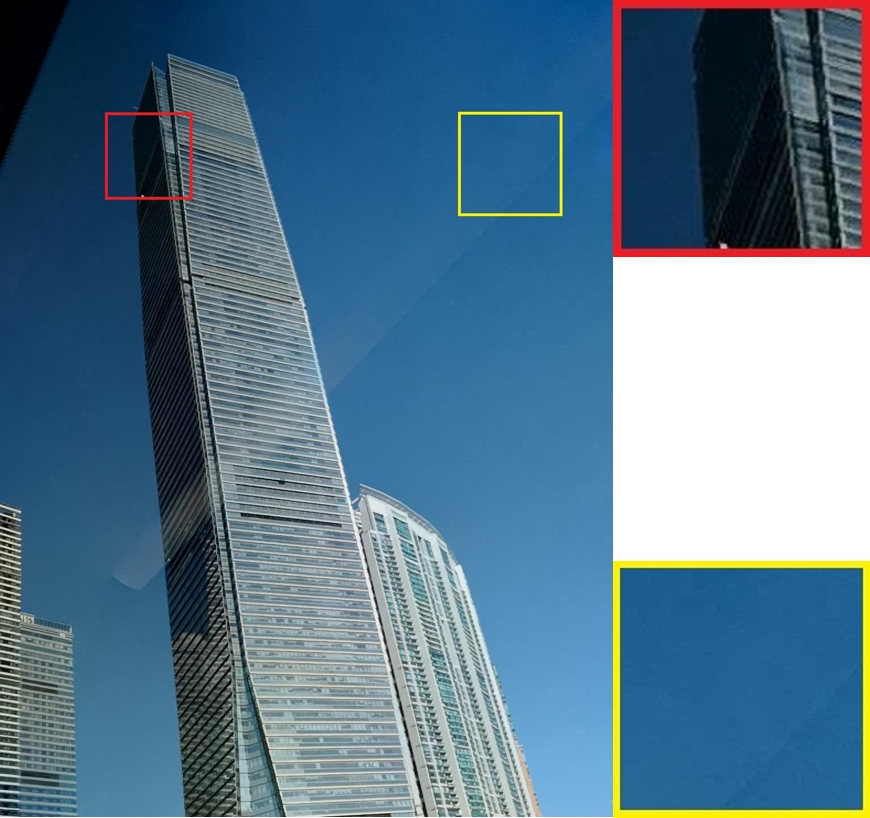}
        \caption{\cite{li2014single}\\~}
        \label{fig:tiger}
    \end{subfigure}
    ~ 
    \begin{subfigure}[b]{0.23\textwidth}
        \includegraphics[width=\textwidth]{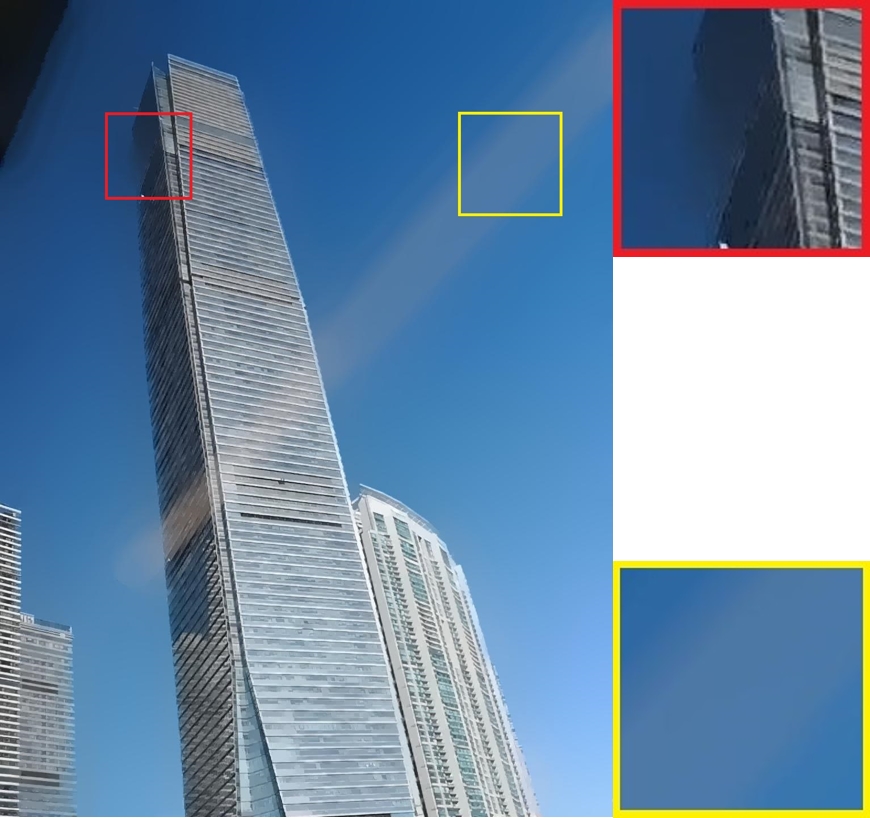}
        \caption{\cite{arvanitopoulos2017single}, \\$\lambda = 0.01$}
        \label{fig:hongkongL0}
    \end{subfigure}
    ~
    \begin{subfigure}[b]{0.23\textwidth}
        \includegraphics[width=\textwidth]{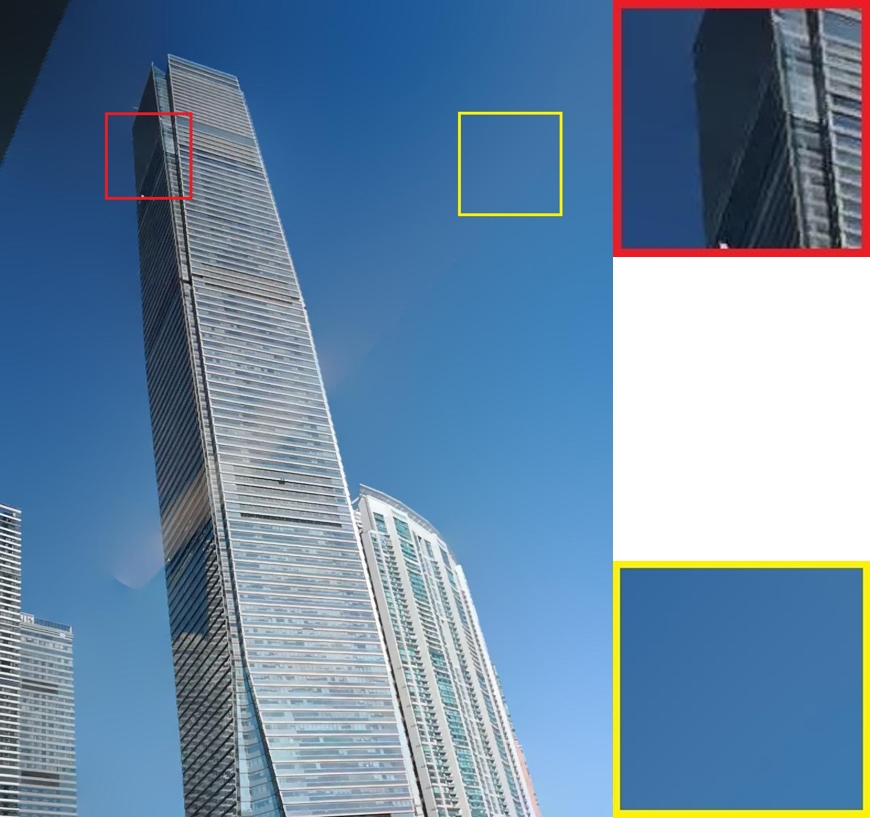}
        \caption{Proposed, \\\centering $h = 0.04$}
        \label{fig:mouse}
    \end{subfigure} \\\vspace{3mm}
    
    \begin{subfigure}[b]{0.23\textwidth}
        \includegraphics[width=\textwidth]{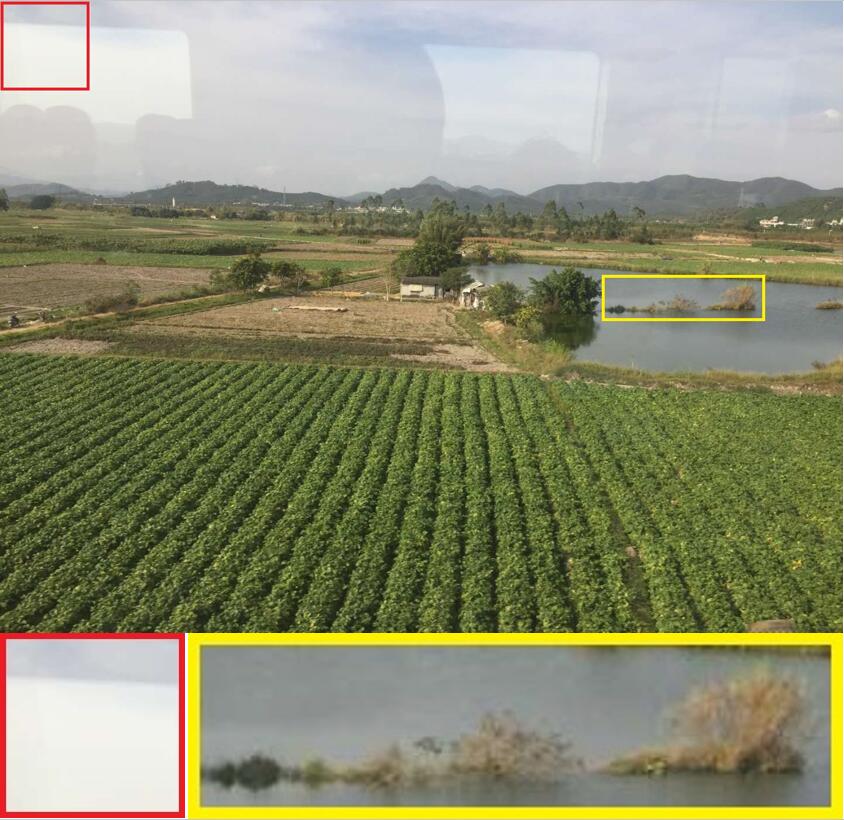}
        \caption{Input 2\\~}
        \label{fig:mouse}
    \end{subfigure}
    ~
    \begin{subfigure}[b]{0.23\textwidth}
        \includegraphics[width=\textwidth]{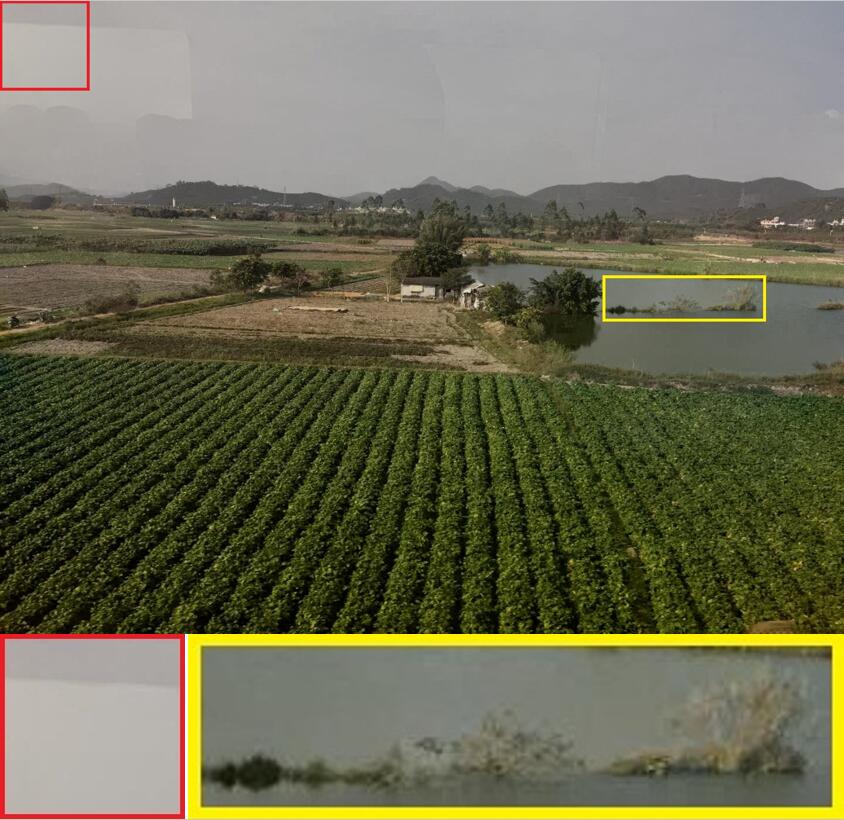}
        \caption{\cite{li2014single}\\~}
        \label{fig:fieldLi}
    \end{subfigure}
    ~
    \begin{subfigure}[b]{0.23\textwidth}
        \includegraphics[width=\textwidth]{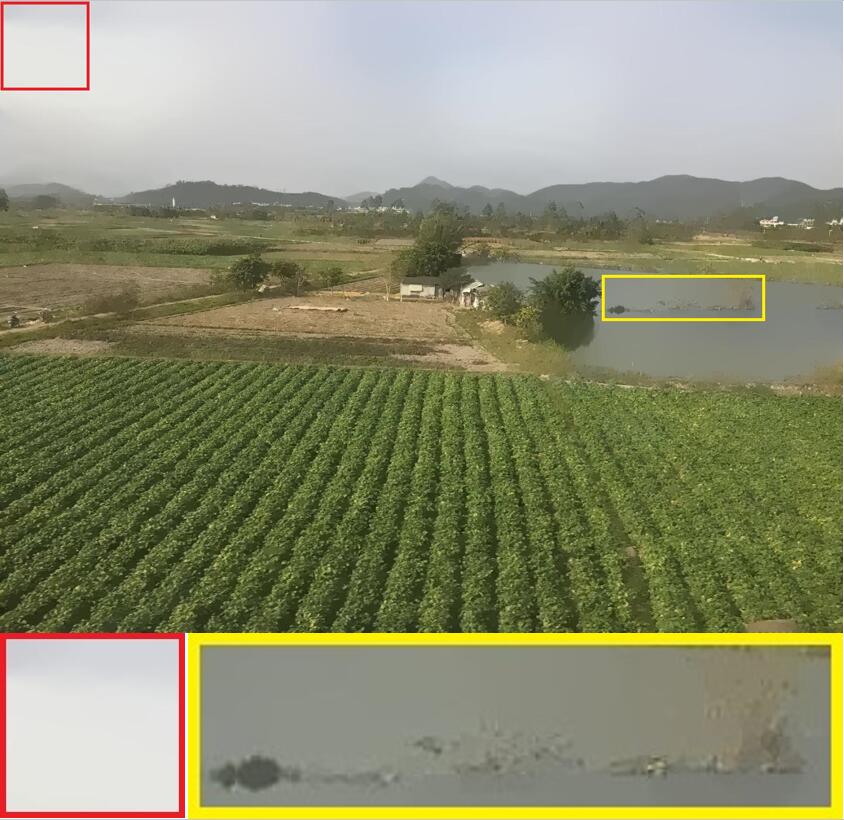}
        \caption{\cite{arvanitopoulos2017single}, \\$\lambda = 0.005$}
        \label{fig:fieldL0}
    \end{subfigure}
    ~
    \begin{subfigure}[b]{0.23\textwidth}
        \includegraphics[width=\textwidth]{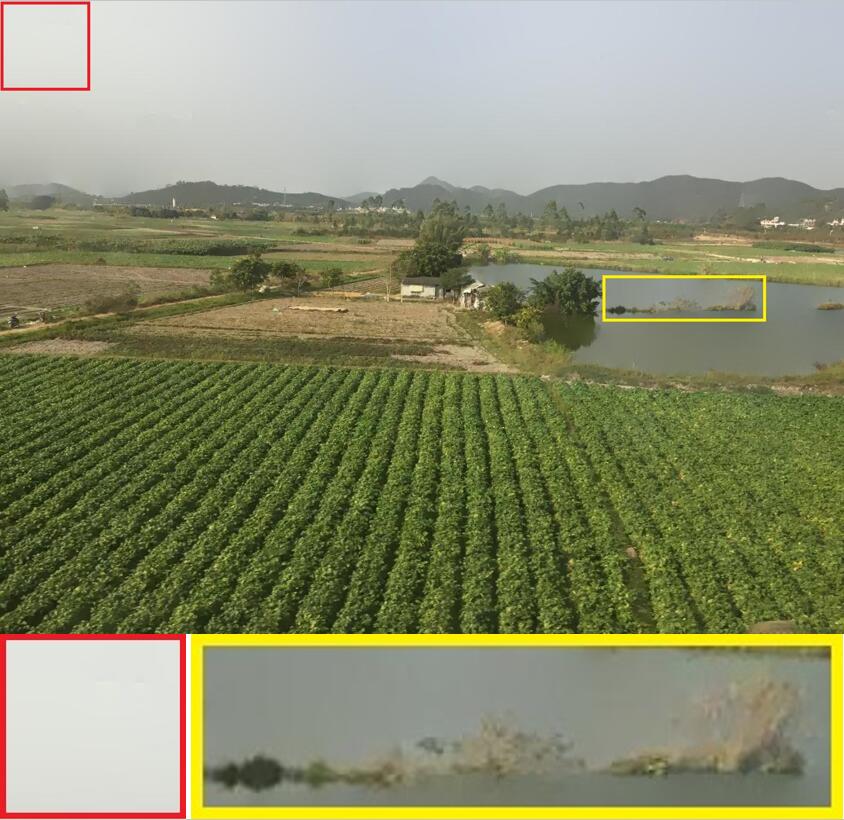}
        \caption{Proposed, \\\centering $h = 0.033$}
        \label{fig:mouse}
    \end{subfigure}\\\vspace{3mm}
    
    \begin{subfigure}[b]{0.23\textwidth}
        \includegraphics[width=\textwidth]{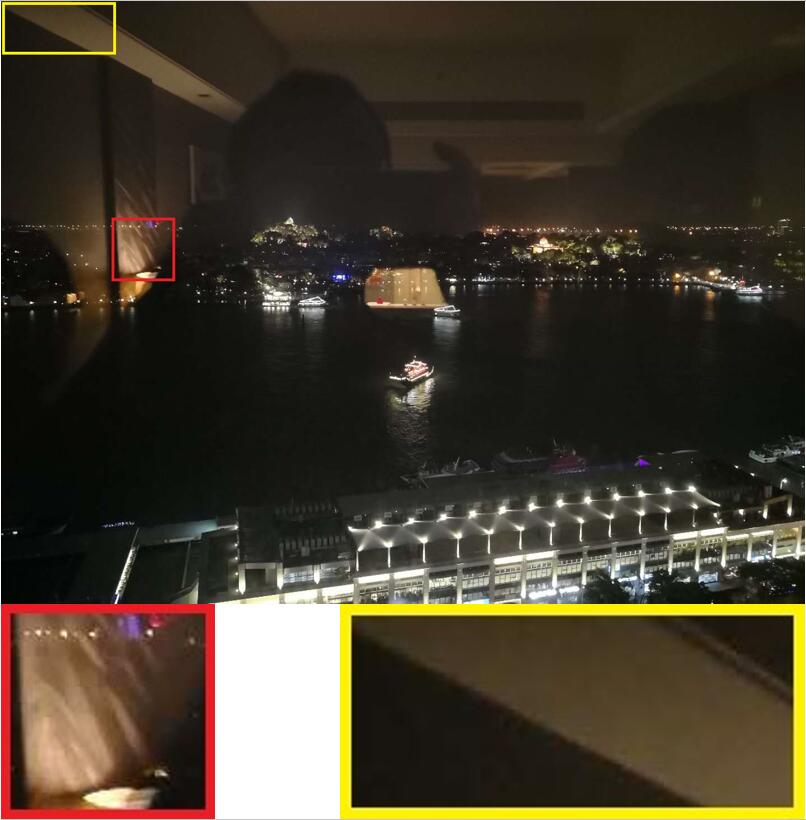}
        \caption{Input 3\\~}
        \label{fig:mouse}
    \end{subfigure}
    ~
    \begin{subfigure}[b]{0.23\textwidth}
        \includegraphics[width=\textwidth]{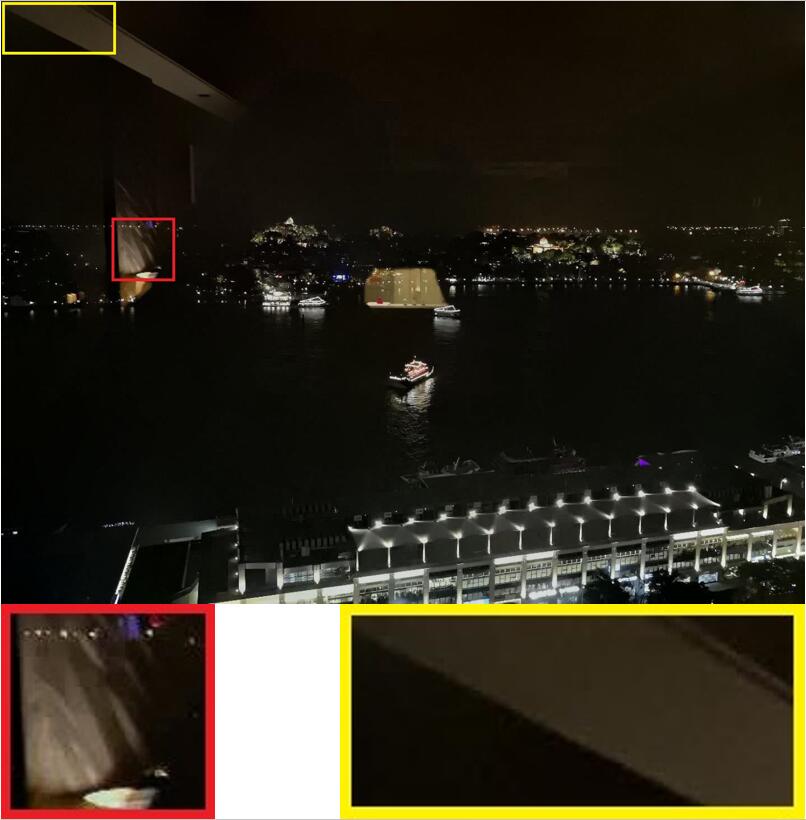}
        \caption{\cite{li2014single}\\~}
        \label{fig:swissgrandLi}
    \end{subfigure}
    ~
    \begin{subfigure}[b]{0.23\textwidth}
        \includegraphics[width=\textwidth]{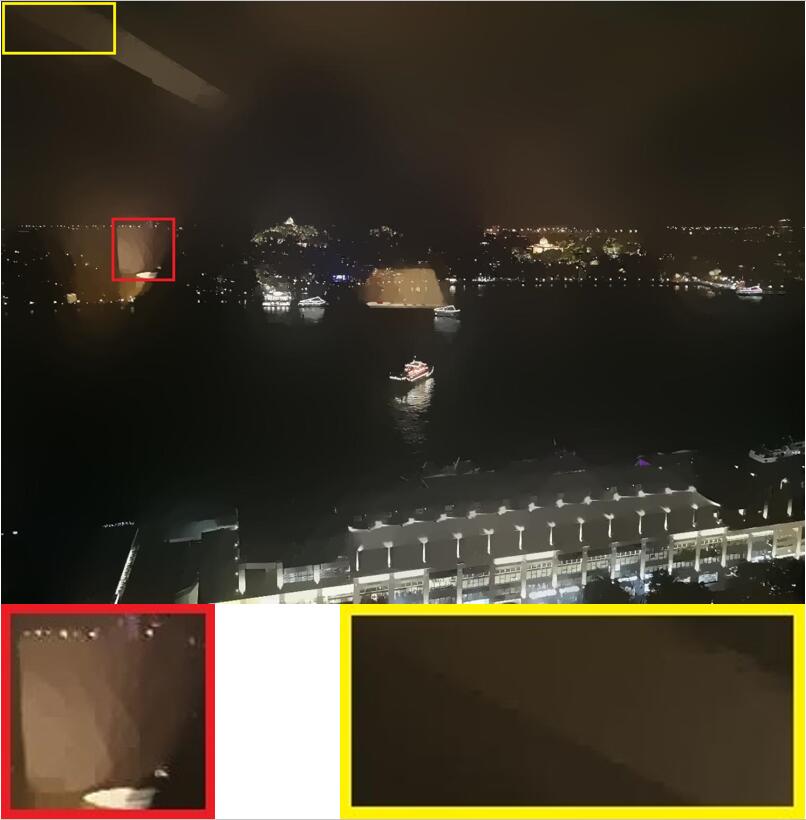}
        \caption{\cite{arvanitopoulos2017single}, \\$\lambda = 0.01$}
        \label{fig:mouse}
    \end{subfigure}
    ~
    \begin{subfigure}[b]{0.23\textwidth}
        \includegraphics[width=\textwidth]{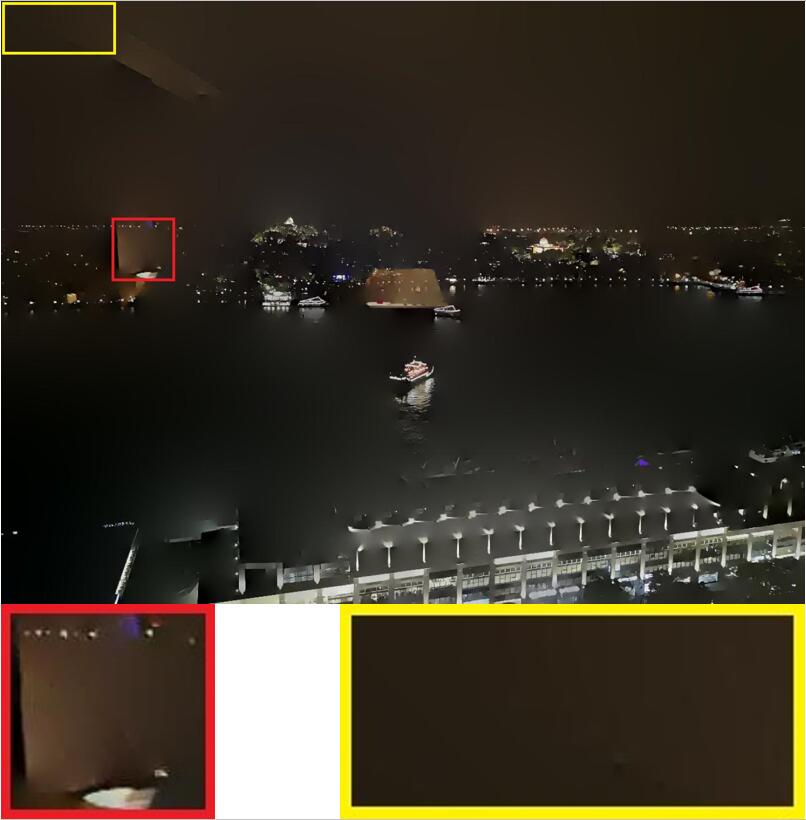}
        \caption{Proposed, \\\centering $h = 0.1$}
        \label{fig:mouse}
    \end{subfigure} \\\vspace{3mm} 
    
    \begin{subfigure}[b]{0.23\textwidth}
        \includegraphics[width=\textwidth]{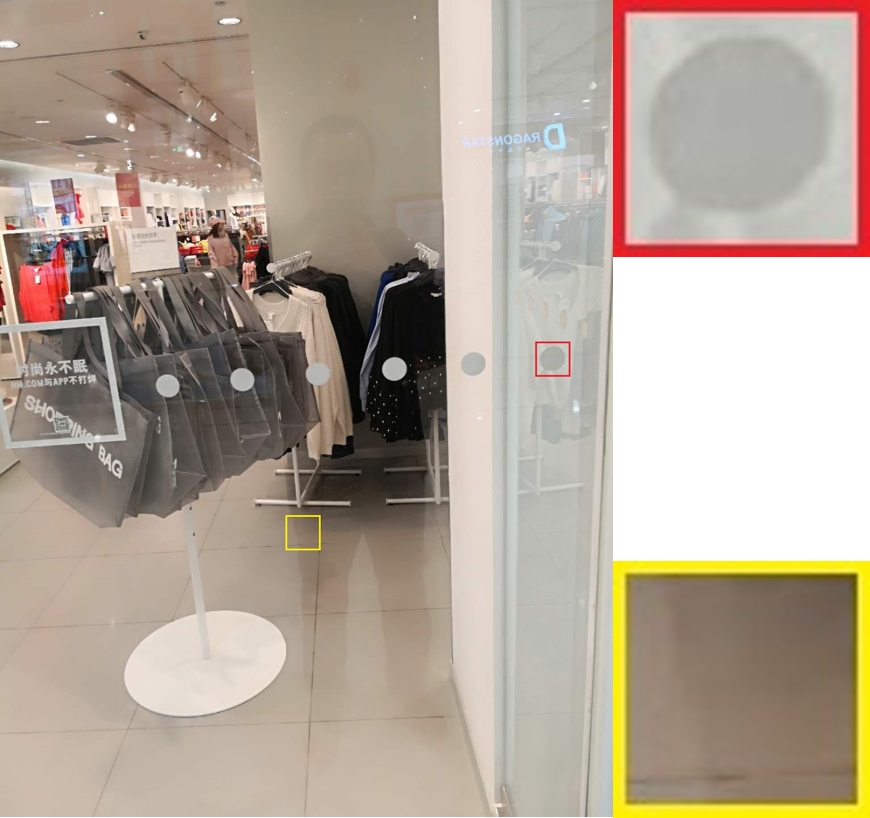}
        \caption{Input 4\\~}
        \label{fig:mouse}
    \end{subfigure}
    ~
    \begin{subfigure}[b]{0.23\textwidth}
        \includegraphics[width=\textwidth]{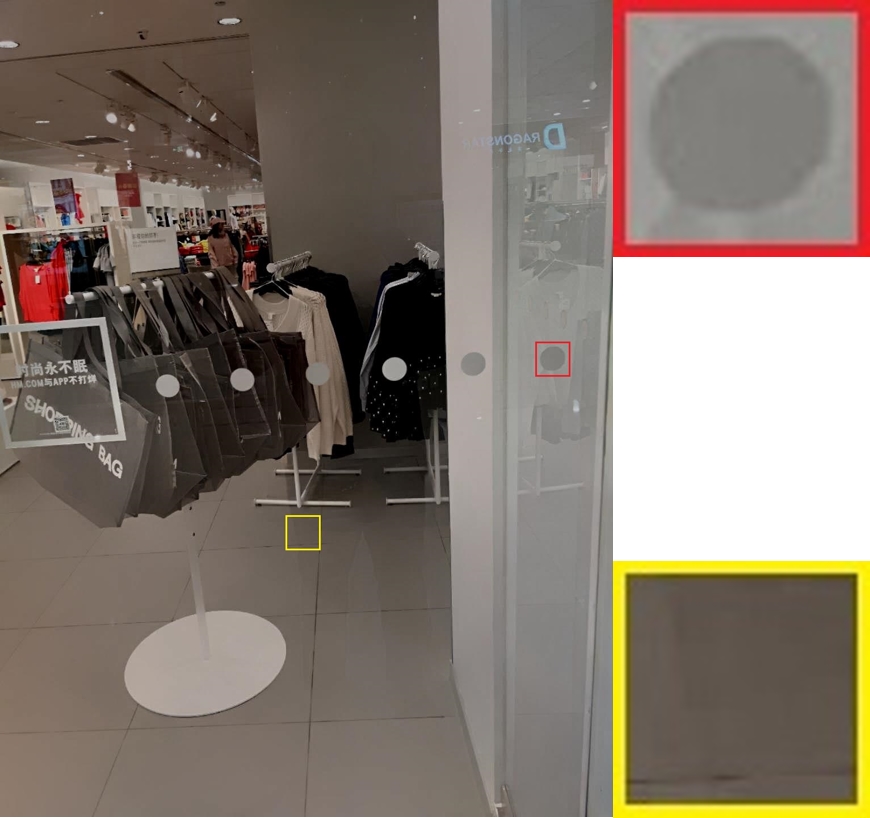}
        \caption{\cite{li2014single}\\~}
        \label{fig:mouse}
    \end{subfigure}
    ~
    \begin{subfigure}[b]{0.23\textwidth}
        \includegraphics[width=\textwidth]{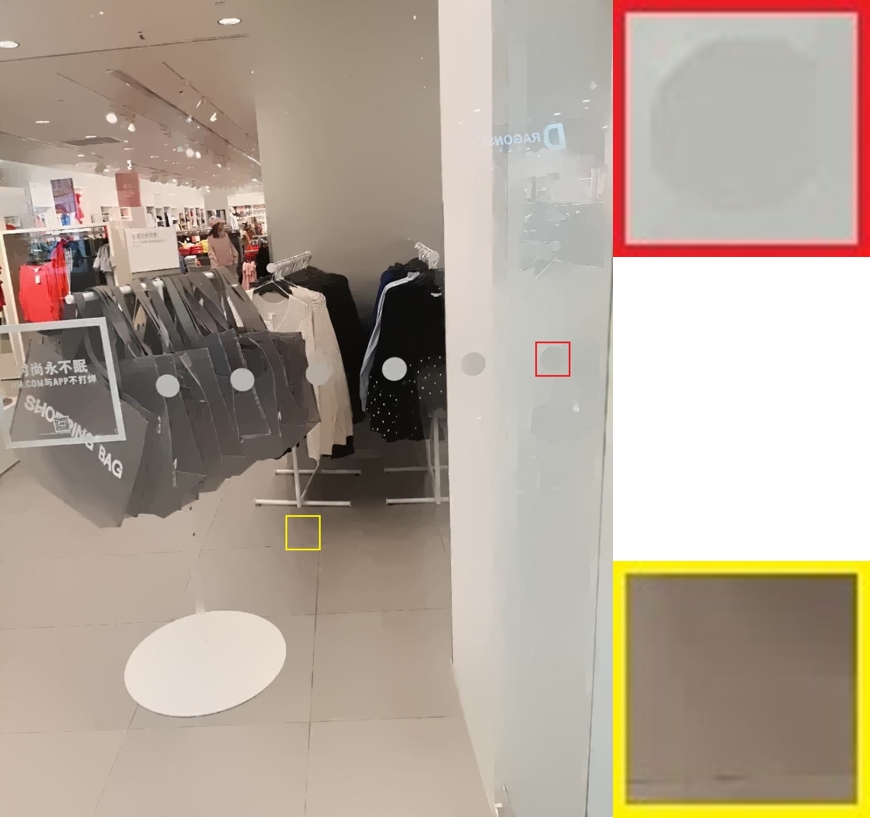}
        \caption{\cite{arvanitopoulos2017single},\\$\lambda = 0.002$}
        \label{fig:xianL0}
    \end{subfigure}
    ~
    \begin{subfigure}[b]{0.23\textwidth}
        \includegraphics[width=\textwidth]{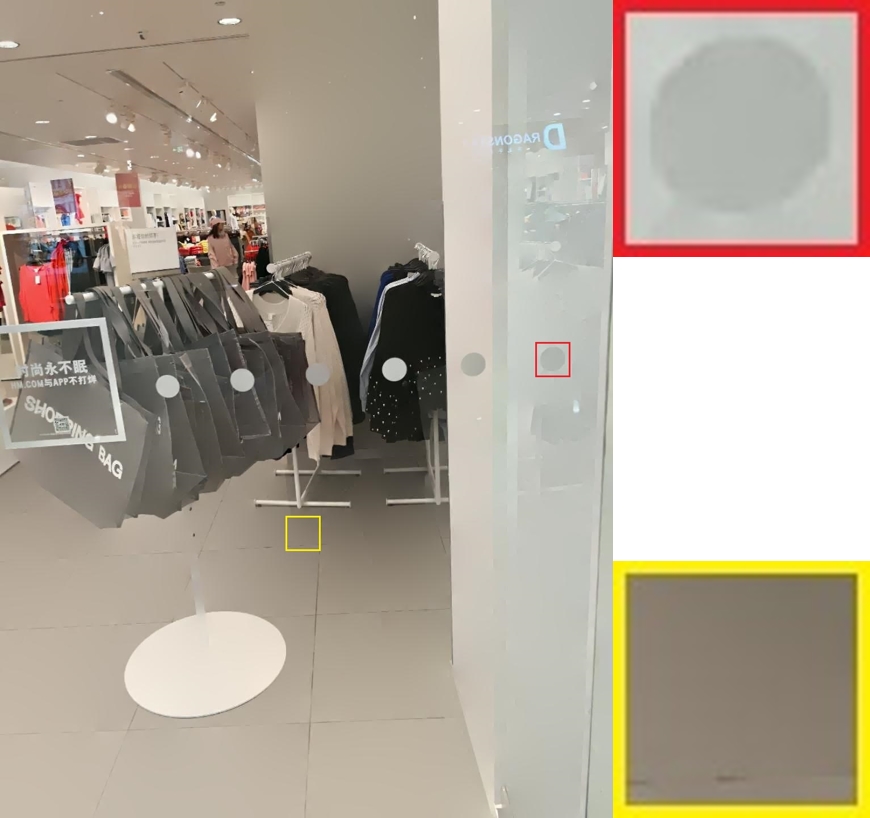}
        \caption{Proposed, \\\centering $h = 0.03$}
        \label{fig:mouse}
    \end{subfigure} 
    
    \caption{Comparison of reflection suppression methods on real-world images taken at various scenes. The method of Li and Brown \cite{li2014single} yields images that appear darker than the original input. Some reflection edges are not completely removed (\eg upper left corner in Fig. \ref{fig:fieldLi} and Fig. \ref{fig:swissgrandLi}). The method of Arvanitopoulos \etal \cite{arvanitopoulos2017single} achieves better color reproduction but suffers from some loss of details in the transmission layer (\eg the top corner of the building in Fig. \ref{fig:hongkongL0}, the vegetation in Fig. \ref{fig:fieldL0}, the disk on the glass in Fig. \ref{fig:xianL0}). Our Proposed method retains the most transmission layer details with superior reflection layer suppression among these methods. Best viewed on screen.}\label{fig:real}
\end{figure*}

\begin{figure*}
\nocaption
\centering
    \begin{subfigure}[b]{0.315\textwidth}
        \includegraphics[width=\textwidth]{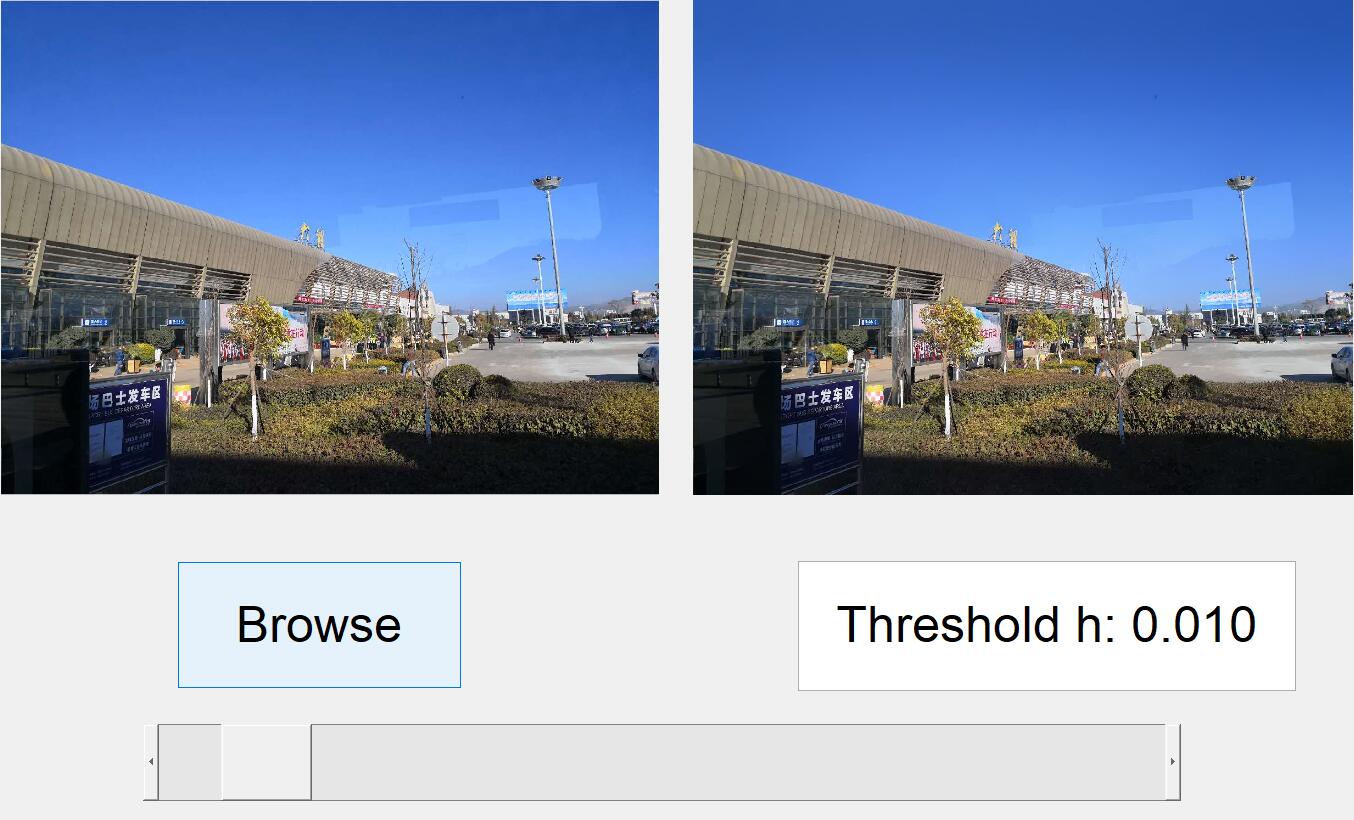}
        \caption{$h = 0.01$}
        \label{fig:gull}
    \end{subfigure}
    ~
    \begin{subfigure}[b]{0.315\textwidth}
        \includegraphics[width=\textwidth]{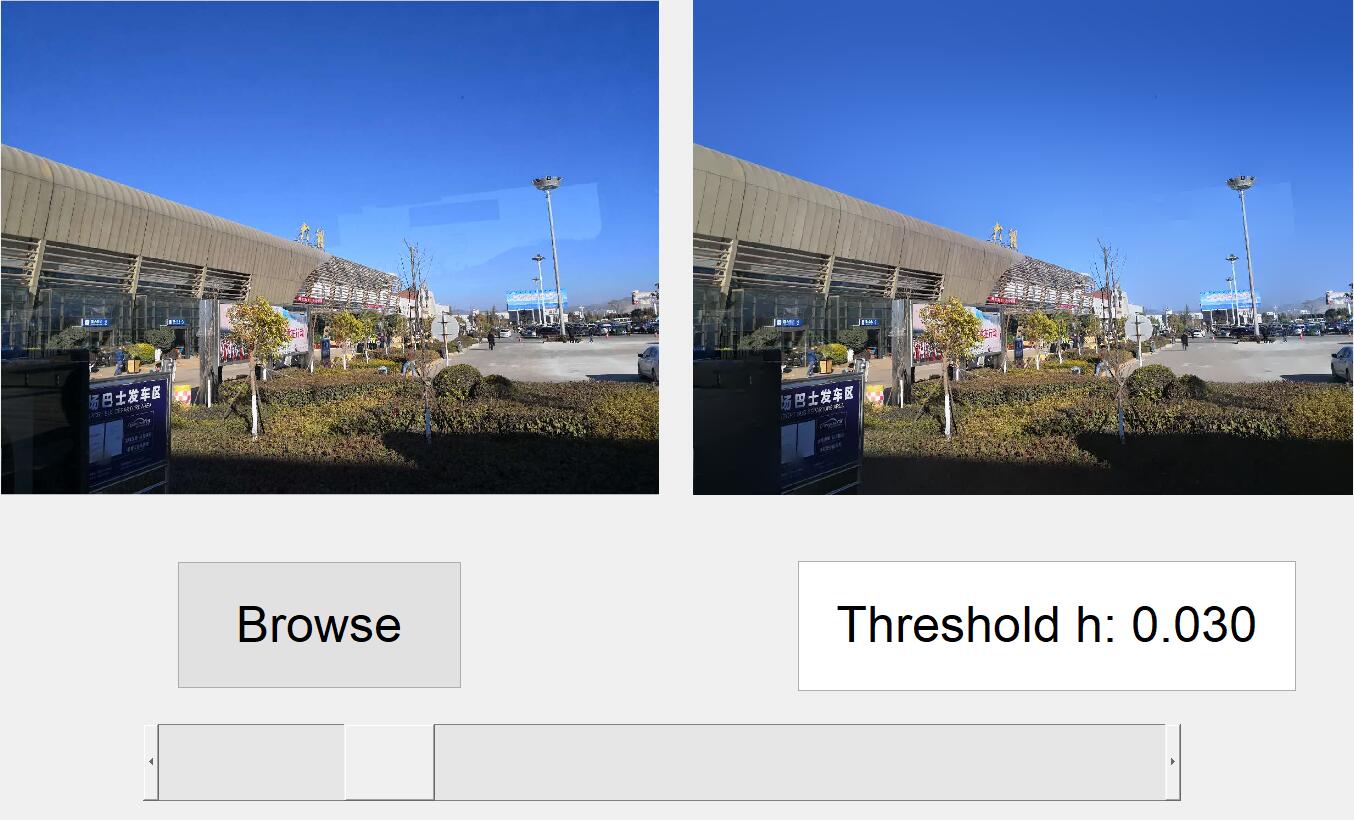}
        \caption{$h = 0.03$}
        \label{fig:tiger}
    \end{subfigure}
    ~
    \begin{subfigure}[b]{0.315\textwidth}
        \includegraphics[width=\textwidth]{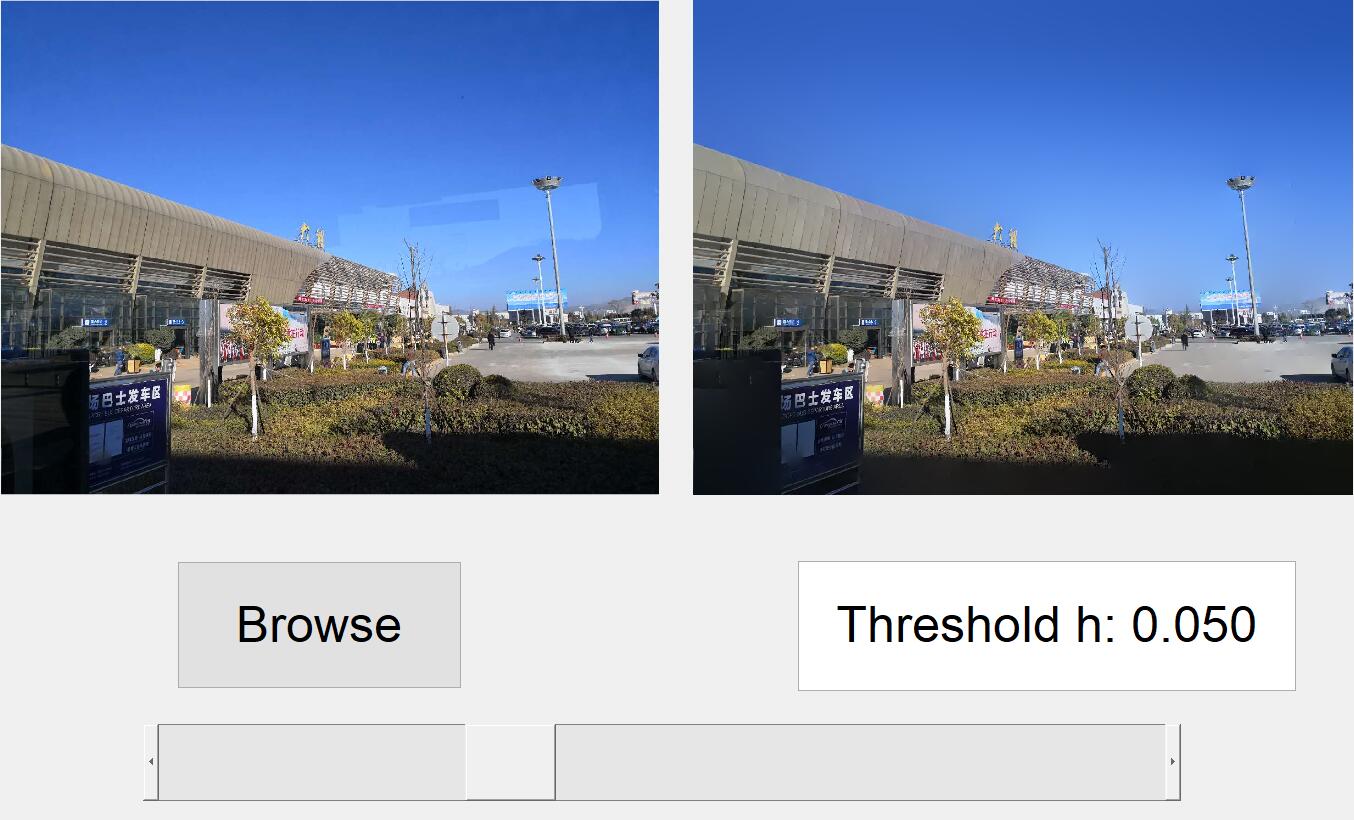}
        \caption{$h = 0.05$}
        \label{fig:hongkongL0}
    \end{subfigure}
    \caption{A slider demo simulated in MATLAB. As we move the slider to the right, the $h$ value increases and the reflection is gradually suppressed. The response time is less than 1.5 seconds for smartphone images of size $1080 \times 1440$. Best viewed on screen.}
    \label{fig:slider}
\end{figure*}

Table \ref{time real} demonstrates the advantage of the proposed model in terms of the execution time. It is much faster compared to other state-of-the-art algorithms.\footnote{At such picture size, the approach in Wan \etal \cite{wan2016depth} reports out-of-memory error, indicating that it is not suitable for large-sized images.} Typically it only takes less than 1.5 seconds to output the dereflected images. Moreover, the dereflection quality also outperforms other methods as demonstrated in Fig.\,\ref{fig:real} (Notice the difference in the zoomed-in boxes). Our proposed method not only suppresses the reflection satisfactorily but also maintains as much transmission details as possible. Being fast and effective, our proposed method has the potential of being implemented directly on portable devices such as smartphones and tablets. The high efficiency makes it possible for a mobile device user to adjust the parameter $h$ easily (for example, via moving a slider on the phone screen) to get an immediate response and select the best dereflected image according to the user's visual perception (See Fig. \ref{fig:slider}). 

\begin{figure*}
\nocaption
\centering
    \begin{subfigure}[b]{0.23\textwidth}
        \includegraphics[width=\textwidth]{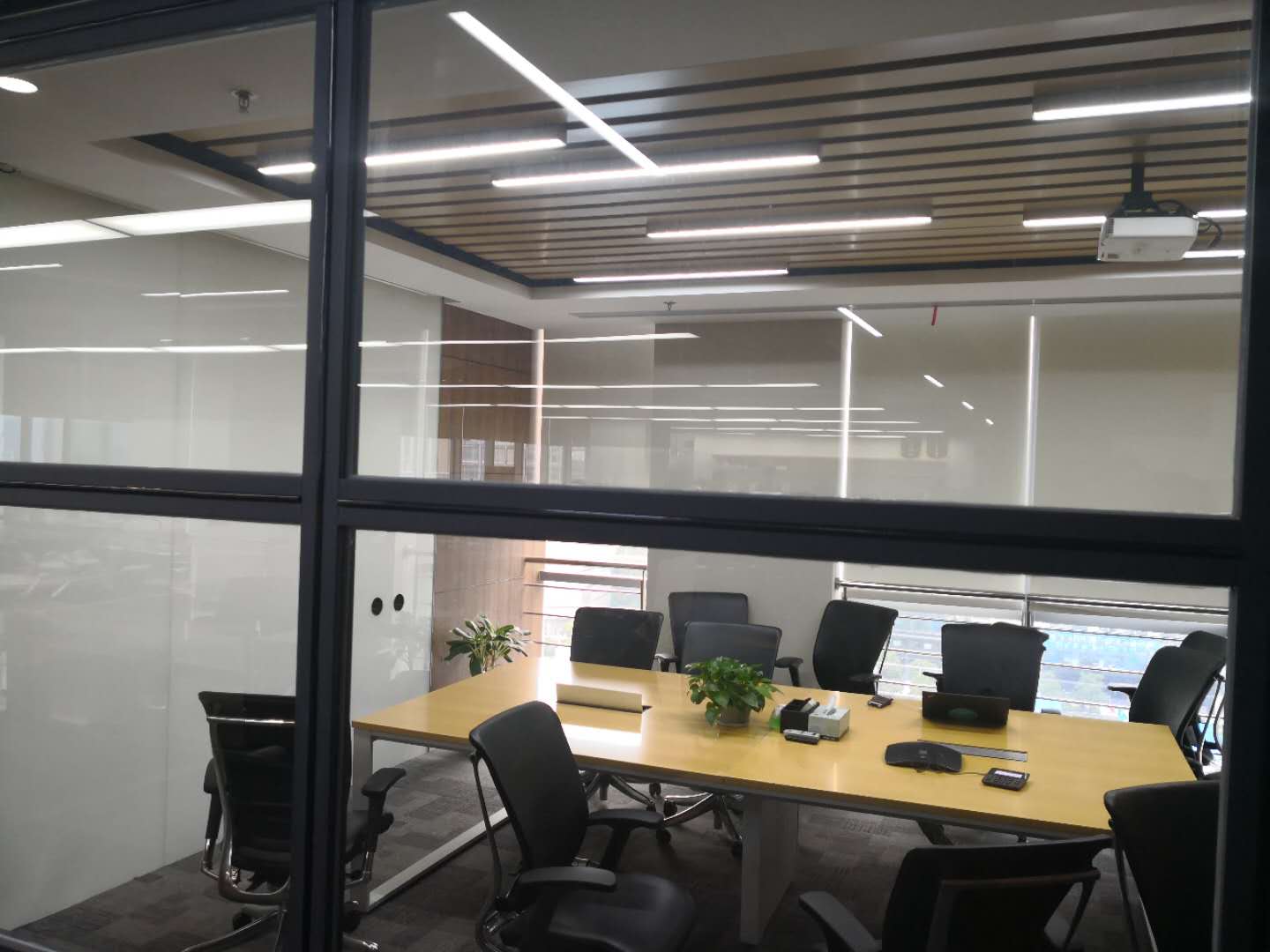}
        \caption{Input 1\\~}
    \end{subfigure}
    ~
    \begin{subfigure}[b]{0.23\textwidth}
        \includegraphics[width=\textwidth]{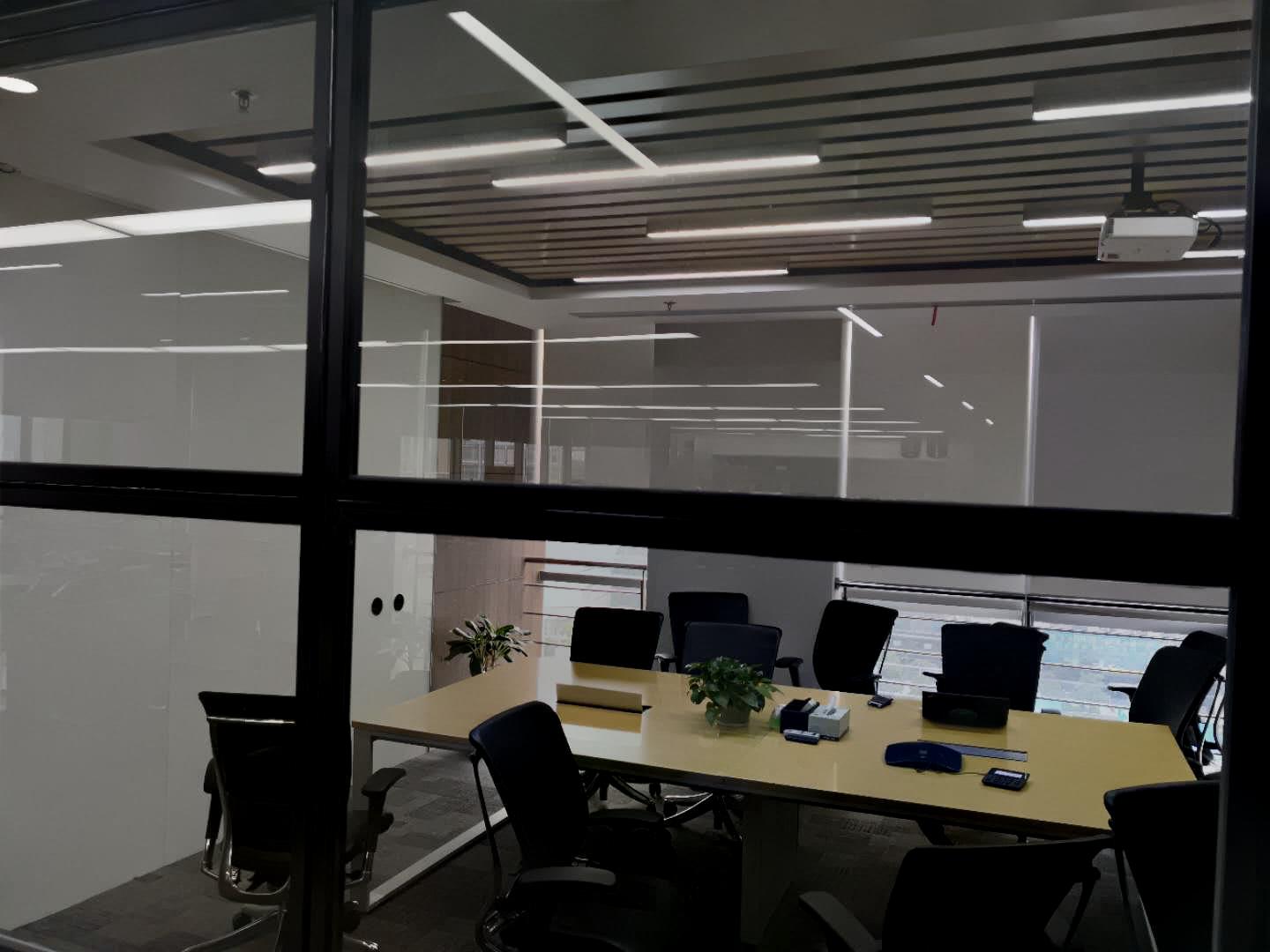}
        \caption{\cite{li2014single}\\~}
    \end{subfigure}
    ~
    \begin{subfigure}[b]{0.23\textwidth}
        \includegraphics[width=\textwidth]{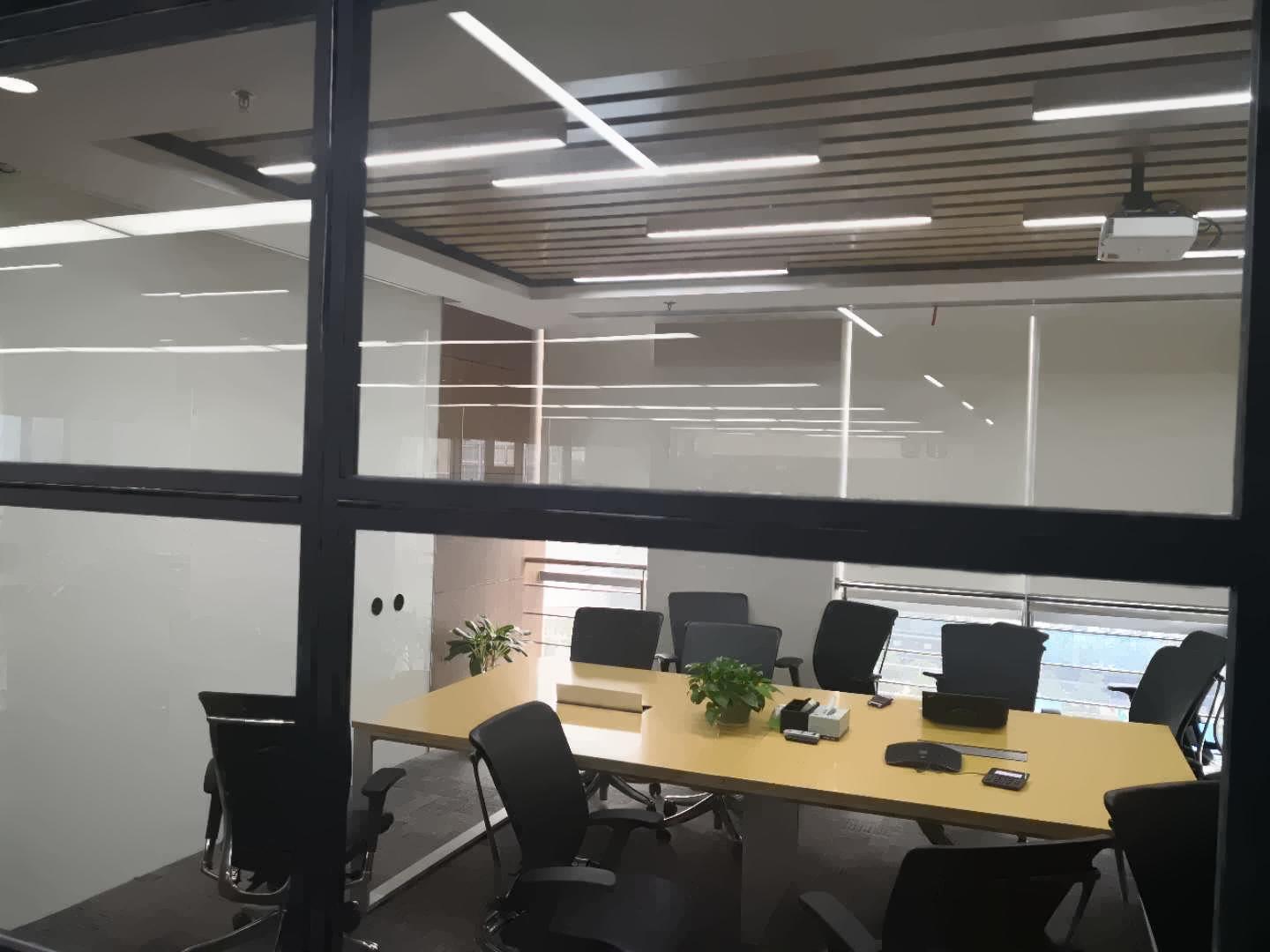}
        \caption{\cite{arvanitopoulos2017single}, \\$\lambda = 0.002$}
    \end{subfigure}
    ~
    \begin{subfigure}[b]{0.23\textwidth}
        \includegraphics[width=\textwidth]{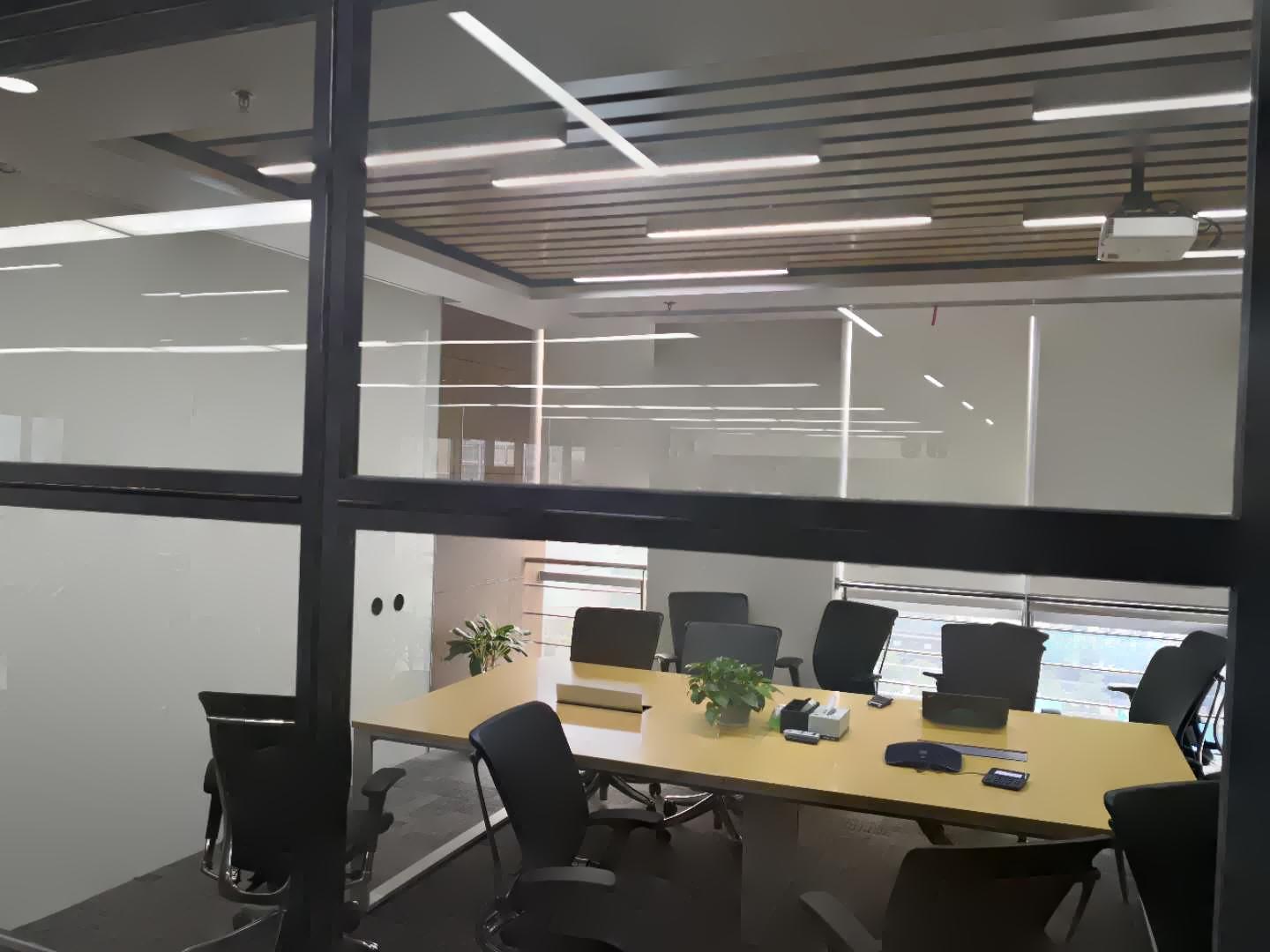}
        \caption{Proposed, \\\centering $h = 0.03$}
    \end{subfigure} \\\vspace{3mm}
    
    \begin{subfigure}[b]{0.23\textwidth}
        \includegraphics[width=\textwidth]{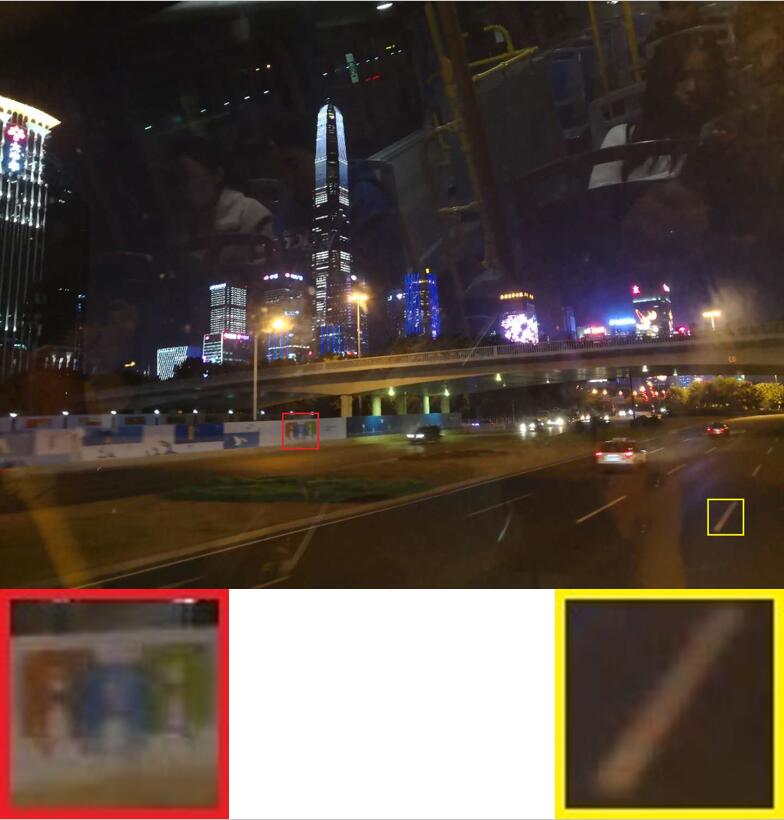}
        \caption{Input 2\\~}
        \label{fig:pinganIFC}
    \end{subfigure}
    ~
    \begin{subfigure}[b]{0.23\textwidth}
        \includegraphics[width=\textwidth]{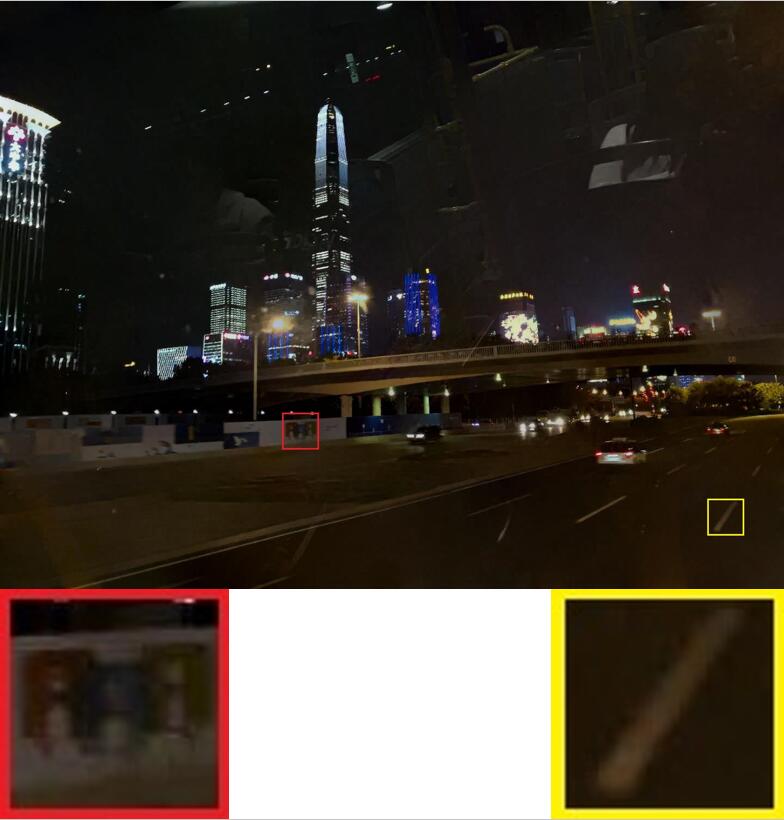}
        \caption{\cite{li2014single}\\~}
    \end{subfigure}
    ~
    \begin{subfigure}[b]{0.23\textwidth}
        \includegraphics[width=\textwidth]{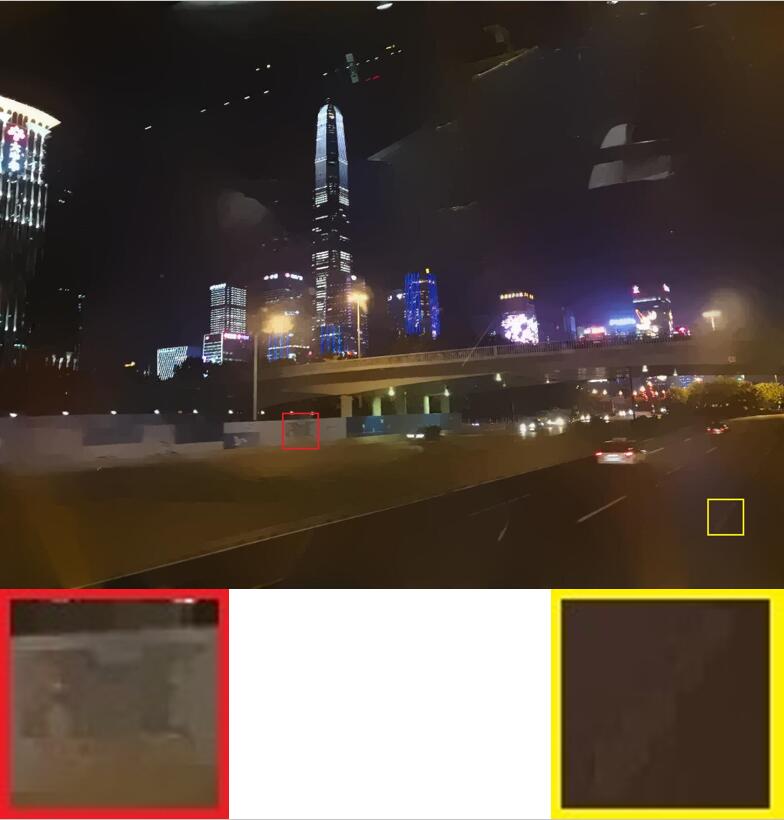}
        \caption{\cite{arvanitopoulos2017single}, \\$\lambda = 0.002$}
    \end{subfigure}
    ~
    \begin{subfigure}[b]{0.23\textwidth}
        \includegraphics[width=\textwidth]{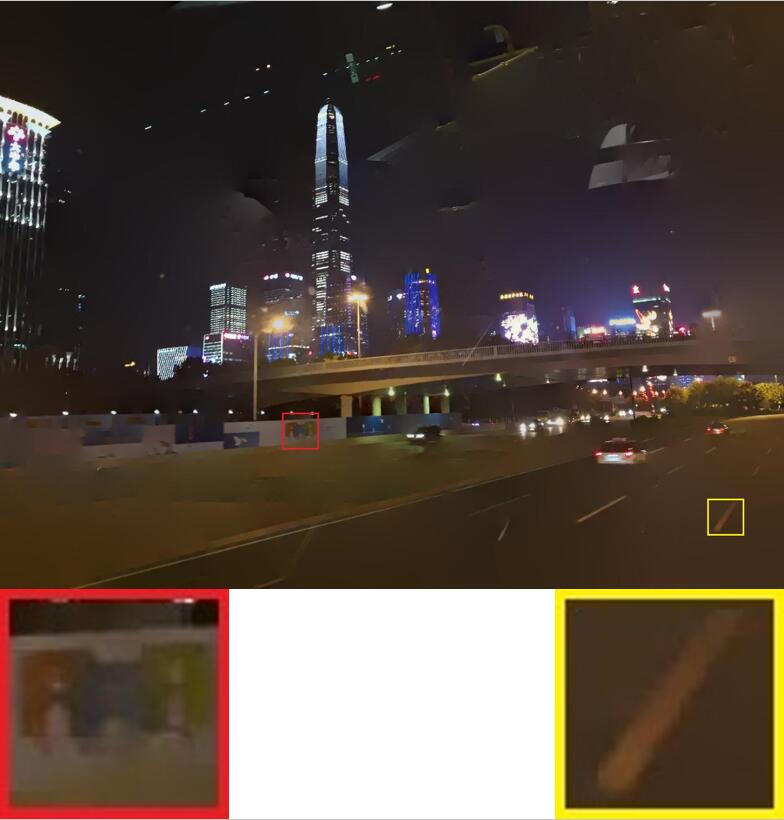}
        \caption{Proposed, \\\centering $h = 0.03$}
    \end{subfigure}
    \caption{Failure cases of our proposed method. Failure is likely to occur when edges in the reflection layer are sharp and strong. This limitation is also observed in the other two methods. In Row 1, the reflection of the fluorescent lamps outside the room is almost as sharp as the real ones inside, which makes it hard to distinguish between them. In Row 2, although our proposed method fails to completely remove the reflection (the inside of a bus), it retains more transmission details than \cite{arvanitopoulos2017single} as shown in the zoomed-in regions. The method in \cite{li2014single} again produces dark outputs. Best viewed on screen.}
    \label{fig:failure}
\end{figure*}

However, our model also has limitation when the model assumption (\ref{eqn:assumption}) is violated. If the reflection layer contains sharp edges, the corresponding gradients at the edge pixels will be large. Therefore, increasing the threshold parameter $h$ won't removed these reflection edges before losing some gentle transmission layer details. Failure cases are shown in Fig.\,\,\ref{fig:failure}, where none of the methods in comparison completely removes the reflection. That being said, our proposed method still retains more details even if edges in the transmission layer are not sharp enough, for example, in dark images like Fig. \ref{fig:pinganIFC}. 

\begin{table*}[t]
\centering
\caption{Comparison of PSNR and SSIM of reflection suppression methods on synthetic images in Fig. \ref{synthetic comparison}. Image size: $512 \times 512$ pixels}
\begin{tabular}{C{1.5cm}C{1.35cm}L{1.35cm}C{1cm}C{1cm}R{1.25cm}C{1.25cm}R{1.2cm}C{1.2cm}}
\hline
Image    & \multicolumn{2}{c}{Li and Brown \cite{li2014single}} & \multicolumn{2}{c}{Wan \etal\cite{wan2016depth}} & \multicolumn{2}{c}{Arvanitopoulos \etal \cite{arvanitopoulos2017single}} & \multicolumn{2}{c}{Proposed} \\
\hline
         & PSNR  & SSIM  & PSNR  & SSIM  & PSNR  & SSIM  &     PSNR       &     SSIM       \\
\hline
Fig. \ref{fig:lena_baboon_07}  & 16.08 & 0.549 & 19.81 & 0.874 & 20.87\, & 0.896 & \textbf{20.99} & \textbf{0.903} \\
Fig. \ref{fig:lena_baboon_05}  & 13.46 & 0.344 & 16.65 & 0.700 & 16.80\, & 0.716 & \textbf{16.93} & \textbf{0.736} \\
Fig. \ref{fig:peppers_fruits_07}  & 16.64 & 0.762 & 17.10 & 0.840 & 19.42\, & 0.896 & \textbf{19.44} & \textbf{0.897} \\
Fig. \ref{fig:peppers_fruits_05}  & 13.55 & 0.574 & 14.54 & 0.751 & 15.10\, & 0.787 & \textbf{15.14} & \textbf{0.789} \\
\hline
\end{tabular}
\label{PSNRSSIM}
\vspace{-0.45cm}
\end{table*}

\section{Conclusion and Future Work}\label{sec:conclusion}
We proposed an efficient approach for single image reflection suppression. It is formulated as a convex problem, which is solved via gradient thresholding and solving a variation of 2D Poisson's equation using DCT. We validated the effectiveness and efficiency of our approach through experiments on synthetic and real-world images. It is able to output desirable dereflected smartphone images in seconds. However, single image reflection suppression remains a challenging problem as there are still cases where current approaches fail to completely remove the reflection. Future work includes designing effective and efficient algorithms to handle sharp and strong reflections for large images.

\clearpage
\begin{table*}[t]
\centering
\makebox[0pt][c]{\parbox{\textwidth}{
\begin{minipage}[b]{0.46\textwidth}\centering
\caption{Execution times (\textit{sec}) of reflection suppression methods on synthetic images in Fig. \ref{synthetic comparison}. Image size: $512 \times 512$ pixels}
\begin{tabular}{C{1.5cm}C{1cm}C{1cm}C{1cm}C{1.5cm}}
\hline
Image    & \cite{li2014single} & \cite{wan2016depth} &  \cite{arvanitopoulos2017single} & Proposed \\
\hline
Fig. \ref{fig:lena_baboon_07}&   12.06 & 49.31 & 185.32 & \textbf{0.19} \\
Fig. \ref{fig:lena_baboon_05}&   11.68 & 49.24 & 185.82 & \textbf{0.18} \\ 
Fig. \ref{fig:peppers_fruits_07}& ~\,7.25 & 48.74 & 185.51 & \textbf{0.19} 
\\
Fig. \ref{fig:peppers_fruits_05}& ~\,7.69 & 47.86 & 185.83 & \textbf{0.19} 
\\
\hline
\end{tabular}
\label{time synthetic}
\end{minipage}
\hfill
\begin{minipage}[b]{0.47\textwidth}\flushleft
\caption{Execution times (\textit{sec}) of reflection suppression methods on real-world images in Fig. \ref{fig:real}. Image size: $1080 \times 1440$ pixels}
\begin{tabular}{C{1cm}C{2cm}C{2cm}C{1.5cm}}
\hline
Image    & \cite{li2014single} & \cite{arvanitopoulos2017single} & Proposed \\
\hline
Input 1    & 39.06   & 1044.28 & \textbf{1.46} \\
Input 2    & 52.60   & 1086.05 & \textbf{1.36} \\
Input 3    & 17.90   & 1032.28 & \textbf{1.40} \\
Input 4    & 10.75   & 1100.73 & \textbf{1.15} \\
\hline
\end{tabular}
\label{time real}
\end{minipage}
}
}
\end{table*}

\clearpage

{\small
\bibliographystyle{ieee}
\bibliography{egbib}
}

\end{document}